\pdfoutput=1

\documentclass[11pt]{article}

\usepackage[]{acl}

\usepackage{times}
\usepackage{latexsym}

\usepackage[T1]{fontenc}

\usepackage[utf8]{inputenc}

\usepackage{microtype}

\usepackage{inconsolata}

\usepackage{color}
\usepackage{url}
\usepackage[T1]{fontenc}
\usepackage{multirow}
\usepackage{xtab,afterpage}
\usepackage{multicol}
\usepackage{graphicx}
\usepackage{comment}
\usepackage{float}
\usepackage[export]{adjustbox}
\usepackage{subcaption}
\usepackage{xspace,mfirstuc,tabulary}
\usepackage{todonotes}
\usepackage{booktabs}
\usepackage[utf8]{inputenc}
\usepackage[arabic,farsi,english]{babel}
\usepackage{ragged2e}
\usepackage[T1]{fontenc}
\usepackage{listings}

\definecolor{codegreen}{rgb}{0,0.6,0}
\definecolor{codegray}{rgb}{0.5,0.5,0.5}
\definecolor{codepurple}{rgb}{0.58,0,0.82}
\definecolor{backcolour}{rgb}{0.95,0.95,0.92}

\definecolor{green}{RGB}{0, 153, 0}

\lstdefinestyle{mystyle}{
  backgroundcolor=\color{backcolour}, commentstyle=\color{codegreen},
  keywordstyle=\color{magenta},
  numberstyle=\tiny\color{codegray},
  stringstyle=\color{codepurple},
  basicstyle=\ttfamily\footnotesize,
  breakatwhitespace=false,         
  breaklines=true,                 
  captionpos=b,                    
  keepspaces=true,                 
  numbersep=5pt,                  
  showspaces=false,                
  showstringspaces=false,
  showtabs=false,                  
  tabsize=2
}

\definecolor{chatGPT}{RGB}{255, 128, 0}

\title{Tell Me Why: Explainable Public Health Fact-Checking \\with Large Language Models}

\author{
  Majid Zarharan$^\diamond$, 
  Pascal Wullschleger$^{\diamond,\bullet}$,
  Babak Behkam Kia$^\dagger$\\
  \bf{Mohammad Taher Pilehvar$^\spadesuit$, Jennifer Foster$^\diamond$}
  \ \\
  $^\diamond$ School of Computing, Dublin City University
  \\
  $^\dagger$
  Iran University of Science and Technology
   \ \\
  $^\spadesuit$
  School of Computer Science and Informatics, Cardiff University
  \ \\
  $^\bullet$
  Lucerne School of Computer Science and Information Technology
  \\
  \texttt{majid.zarharan2@mail.dcu.ie}
  }

\date{}

\begin{document}
\maketitle
\begin{abstract}
This paper presents a comprehensive analysis of explainable fact-checking through a series of experiments, focusing on the ability of large language models to verify public health claims and provide explanations or justifications for their veracity assessments.
We examine the effectiveness of zero/few-shot prompting and parameter-efficient fine-tuning across various open and closed-source models, examining their performance in both isolated and joint tasks of veracity prediction and explanation generation. 
Importantly, we employ a dual evaluation approach comprising previously established automatic metrics and a novel set of criteria through human evaluation. 
Our automatic evaluation indicates that, within the zero-shot scenario, GPT-4 emerges as the standout performer, but in few-shot and parameter-efficient fine-tuning contexts, open-source models demonstrate their capacity to not only bridge the performance gap but, in some instances, surpass GPT-4.
Human evaluation reveals yet more nuance as well as indicating potential problems with the gold explanations. 
\end{abstract}

\section{Introduction}
The recent COVID-19 pandemic has highlighted the critical need for fact-checking within the public health domain. 
In an era where information spreads swiftly across social media platforms, the feasibility of manual fact-checking is significantly challenged. 
Misinformation within the health domain can have severe, even fatal consequences, underscoring the vital role of automated fact-checking mechanisms in averting potential crises and protecting public health \cite{Public-Health, sarrouti-etal, healthfc}.

The ability to provide clear explanations is a crucial part of effective fact-checking, given that fact-checkers need to convince their audience of their 
evidence-backed 
conclusions \cite{Survey-Fact-Checking}.
While certain machine learning models like decision trees and linear regression inherently offer a degree of explainability due to their simple operational frameworks, the landscape changes drastically with neural network-based Large Language Models (LLMs). 
These models, which stand at the cutting edge of automated fact-checking, present significant challenges in terms of interpretability and explainability \cite{Generating-Fact-Checking-Explanations-atanasova}.
To address these challenges, there have been efforts to develop explainable fact-checking methods that employ attention mechanisms, rule discovery, or summarization techniques \cite{Explainable-Fact-Checking-Survey}. 
Our study focuses on Natural Language Explanation (NLE), a strategy where models generate textual justifications for their predictions tailored to specific inputs.

\begin{table*}[ht!]
    \small
    {
    \begin{tabular}{p{15.5cm}}
        \toprule
        \multicolumn{1}{c}{\textbf{Context}} \\
        \midrule
        The Pennsylvania Department of Health says people may have been exposed to measles between Aug. 22 and Aug. 29 in York County and Hershey. Health officials say a patient in WellSpan York Hospital has a confirmed case of measles, which can be highly contagious. The hospital is notifying patients, staff and visitors who were in either the hospital or WellSpan Stony Brook Health Center. Officials say the risk of getting measles is minimal for anyone properly immunized against the disease.\\
        
        \end{tabular}

    \begin{tabular}{p{5.2cm}p{1cm}p{8.5cm}}
        \midrule
        \multicolumn{1}{c}{\textbf{Claim}} & \multicolumn{1}{c}{\textbf{Label}} & \multicolumn{1}{c}{\textbf{Explanation}}\\ 
         \midrule
         Public warned of possible measles exposure in Pennsylvania.
         & True
         & State health authorities are warning the public about possible measles exposure at a number of Pennsylvania locations over the past week.\\
        \bottomrule
    \end{tabular}
    }
  \caption{A random sample from PUBHEALTH test set. The context is a summary of the original context.}
  \label{random-sample} 
\end{table*}

To our knowledge, the application of LLMs to the generation of explanations in fact-checking contexts remains unexplored.
Here, we take a step in this direction by carrying out an extensive evaluation of both open- and closed-source LLMs in assessing the veracity of public health claims and in generating explanations for these assessments. We report results for zero- and few-shot prompt-based learning \cite{Pre-Train-Prompt-predict} and Parameter-Efficient Fine-Tuning \cite[PEFT]{peft}.

In assessing the quality of the explanations generated, we employ a dual evaluation strategy that combines automatic metrics with human evaluation.
This holistic approach is designed to capture a more accurate picture of explanation effectiveness, recognizing that a single metric or method may not fully grasp the 
nuances of explanation quality \cite{Local-Interpretations}.

According to our automatic evaluation, the GPT family of LLMs outperform the open-source models (\textit{Falcon-180B, Llama-70b, Vicuna-13, Mistral-7b}) on the task of veracity prediction in the zero-shot setting. This performance gap narrows in the few-shot setting, showcasing the potential of open-source models with limited examples.
The best performance is achieved using PEFT. 
This trend persists across both veracity prediction and explanation generation tasks. Human evaluation demonstrates that GPT-4, in a zero-shot setting, excels in generating explanations that meet various evaluation criteria effectively. 
Further detailed manual analysis of the explanations generated in both isolated and joint tasks reveals that explanations produced in the context of the joint task tend to be of higher quality than those generated for the explanation task alone.

Our contributions are two-fold: 1) we introduce a novel set of guidelines for human evaluation of explainable fact-checking, which we manually apply to hundreds of LLM-generated explanations, yielding new insights.\footnote{\url{https://github.com/Zarharan/NLE-for-fact-checking}} 
2) we conduct an extensive series of experiments on the PUBHEALTH dataset using closed- and open-source state-of-the-art LLMs, exploring their strengths and weaknesses via both human and automatic evaluations.
\section{Related Work}
\paragraph{Fact Checking Datasets.} 
Some fact-checking datasets include explanations that were collected or generated automatically \cite{alhindi-dataset,e-FEVER, ExClaim}. Other datasets \cite{AVeriTeC, Fake-Health} include question-answer pairs for each example to facilitate explainable fact-checking. AVERITEC \cite{AVeriTeC} consists of more than 4.5K real-world claims fact-checked by 50 organizations. Each claim is annotated with question-answer pairs against the open web representing the evidence, a veracity label, and a textual justification describing how the evidence (question-answer pairs) supports the label. The FakeHealth dataset \cite{Fake-Health} introduces binary criteria for use in explainable fake health news detection. \citet{Public-Health} present a novel dataset (PUBHEALTH) for explainable fact-checking in the public health domain. In contrast to the aforementioned datasets, this dataset includes gold explanations by journalists. We use it in our study. A sample is shown in Table~\ref{random-sample}.

\paragraph{Methods.} 
\citet{Generating-Fact-Checking-Explanations-atanasova} and \citet{Public-Health} formulate explanation generation as a summarization task which leads to an extractive explanation.
\citet{Generating-Fact-Checking-Explanations-atanasova} explore veracity prediction, explanation extraction, and a joint model to address both providing explanations and predicting veracity using LIAR-PLUS \cite{alhindi-dataset}. Their joint model achieved the best F1 scores for veracity prediction. However, training jointly with veracity prediction does not outperform the explanation extraction model.

\citet{Explainable-Healthcare-QA} and \citet{Fact-check-Complex-Claims} propose a question-answering (QA) approach to the explanation generation task. %
\citet{Explainable-Healthcare-QA} demonstrate that QA-based methods can be competitive with summarization-based methods, and even more appropriate when relevant information is not explicitly provided. 

\citet{Public-Health} introduce an explanation generation framework based on abstractive-extractive summarization, and propose three different coherence metrics for evaluating the quality of automatically generated explanations. In contrast, we use PUBHEALTH to instruct LLMs to generate an explanation of the claim given a summary of the related context,  focusing on the generation of Natural Language Explanations (NLE) (abstractive) rather than the extractive method. Abstractive methods make explanations flexible \cite{Local-Interpretations} and the models can justify different parts of the context and generate fluent explanations in simpler terms.

\section{Methodology}
\label{Zero-Few-shots-main-section}

\begin{figure}
  \centering
  \includegraphics[scale=.88]{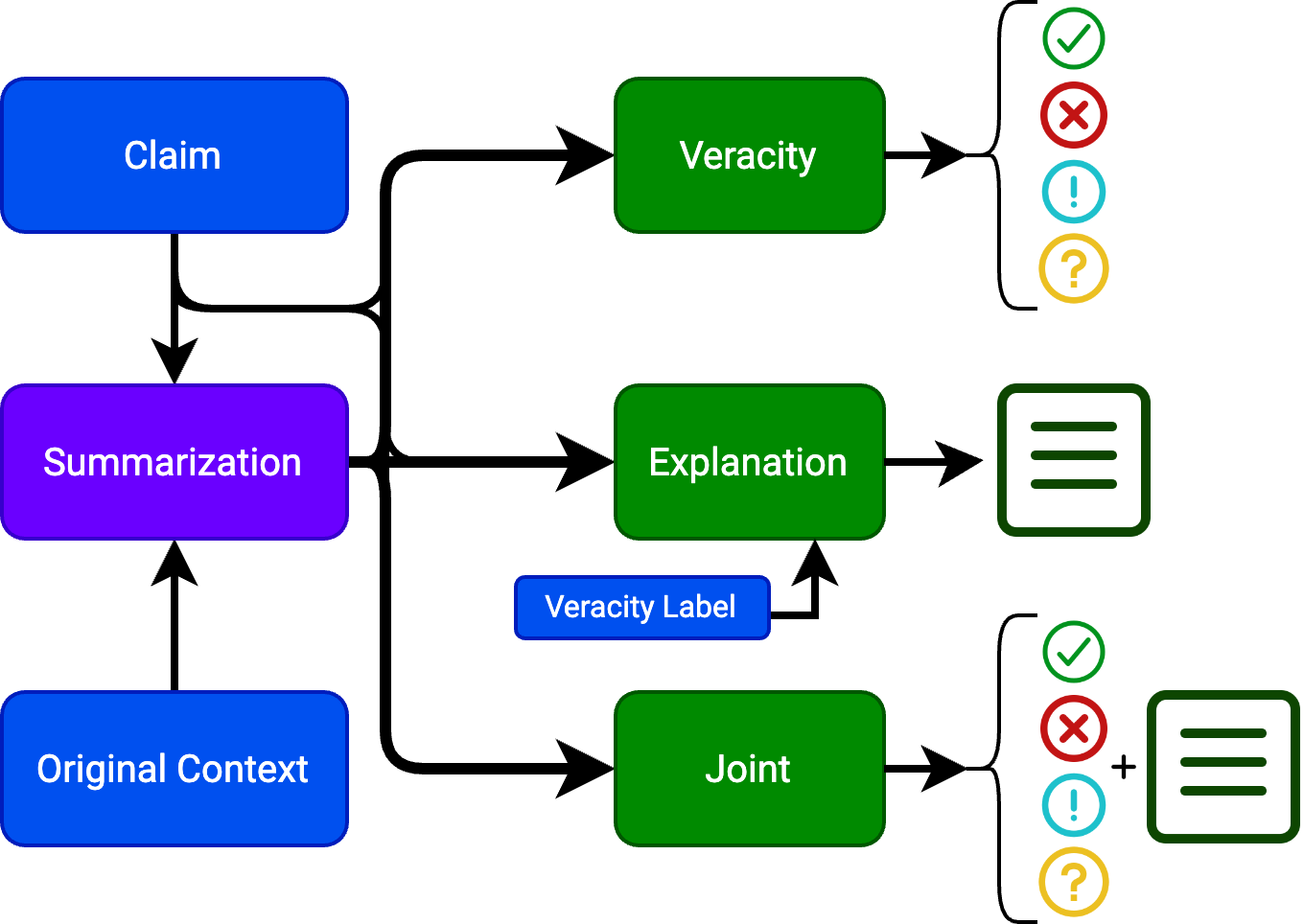}
  \caption{The pipeline for veracity prediction, explanation generation, and the joint setting.}
  \label{pipeline}
\end{figure}

Figure \ref{pipeline} provides a high-level overview of the three tasks considered in our analysis: (1) assessing the veracity of claims, (2) generating corresponding explanations, and (3) joint veracity prediction and explanation. In all tasks, the model receives a summarized version of the original context along with the corresponding claim as inputs. The explanation model is also provided with the gold veracity label.

We instruct various closed- and open-source LLMs with specific prompts for each task. In the few-shot scenario, we individually optimize the number of shots for each model and task. We use the following prompts for the zero-shot experiments, involving closed-source LLMs:

\begin{lstlisting}[linewidth=\columnwidth,language=XML]
%% Veracity prediction
Context: X
Claim: Y
Based only on the context, categorize the claim as:
   - True (supported by context)
   - False (contradicted by context)
   - Mixture (partially supported/contradicted)
   - Unproven (not enough info)
\end{lstlisting}

\begin{lstlisting}[linewidth=\columnwidth,language=XML]
%% Explanation generation
Context: X
Claim: Y
The claim veracity: Z.
Using only the context provided, explain why the claim veracity is Z.
\end{lstlisting}

\begin{lstlisting}[linewidth=\columnwidth,language=XML]
%% Joint model
Context: X
Claim: Y
Based only on the context, categorize the claim as:
   - True (supported by context)
   - False (contradicted by context)
   - Mixture (partially supported/contradicted)
   - Unproven (not enough info)
And explain your reasoning. Provide the response in JSON format with the following keys: veracity, explanation.
\end{lstlisting}


\texttt{X}, \texttt{Y}, and \texttt{Z} respectively represent the context summary, the claim text, and the veracity label of the claim. See Appendix~\ref{appendix:experiment-details} for details of the few-shot tuning process.

Given that the closed-source LLMs are restricted-access models entailing significant costs, we also experiment with open-source LLMs. For fine-tuning, we use parameter-efficient fine-tuning, which aims to reduce the number of trainable parameters and has become a standard paradigm for fine-tuning LLMs \cite{LLM-Survey}. Specifically, we opted for QLoRA \cite{QLoRA} and PEFT \cite{peft}.

\section{Experimental Details}
\subsection{Selected LLMs}

The selected closed-source LLMs include three state-of-the-art LLMs: \textit{GPT-3.5-D} \cite[\texttt{text-davinci-003}]{GPT-3}, \textit{GPT-3.5-T} \cite[\texttt{gpt-3.5-turbo}]{ChatGPT}, and \textit{GPT-4} \cite{gpt4}. We used these models for in-context learning experiments only.
We also use publicly available models, \textit{Falcon-180B} \cite[\texttt{Falcon-180B}]{falcon180}, \textit{Llama-70B} \cite[\texttt{Llama-2-70b}]{llama2}, \textit{Vicuna-13B} \cite[\texttt{vicuna-13b-v1.5-16k}]{vicuna}, and \textit{Mistral-7B} \cite[\texttt{Mistral-7B-v0.1}]{mistral} for in-context learning. Finally, we implement PEFT with \textit{Vicuna-13B} and \textit{Mistral-7B} for all three tasks (see Section~\ref{appendix:settings} of Appendix~\ref{appendix:experiment-details} for more details).

\subsection{Dataset}

We employ PUBHEALTH \cite{Public-Health}, which comprises more than 12.2K claims, each accompanied by journalist-crafted gold-standard explanations (or judgments) to substantiate the fact-check labels assigned to these claims. After collecting data from different fact-checking sources, \citet{Public-Health} preprocessed the data and standardized labels for 4-way classification:  \textit{true}, \textit{false}, \textit{mixture} and \textit{unproven}. Table \ref{LableDistribution} shows the distribution of veracity classes. 

\begin{table}[t!]
    \centering
    \setlength{\tabcolsep}{6pt}
    \resizebox{0.48\textwidth}{!}
    {
    \begin{tabular}{lrrrrr}
        \toprule
        \textbf{Data Split} & \textbf{True} & \textbf{False} & \textbf{Mixture} & \textbf{Unproven} & \textbf{Total} \\ 
         \midrule
         Train & 5,077 & 2,999 & 1,432 & 290 & 9,798 \\
         Val & 629 & 380 & 163 & 41 & 1,213 \\
         Test & 599 & 387 & 201 & 45 & 1,232 \\
        \midrule
        Total & 6,305 & 3,766 & 1,796 & 376 & 12,243 \\
        \bottomrule
    \end{tabular}
    }
  \caption{The distribution of samples in PUBHEALTH across the four veracity labels.
  }
  \label{LableDistribution} 
\end{table}

\subsection{Context Summarization}
In the PUBHEALTH dataset, the mean and median word counts of articles are approximately 700 and 600 words respectively. So, to address the sequence length limitation in different LLMs, particularly in our few-shot experiments, we summarized the context of all instances in the dataset. Following \citet{Benchmarking-llm-Summarization}, who conducted a human evaluation of news summary datasets and discovered that the zero-shot summaries generated by instruction-based LLMs were on par with summaries written by humans, we manually compared the summaries generated by two LLMs 1) \texttt{gpt-3.5-turbo} and 2) \texttt{text-davinci-003}, on a small training set sample. 
We tested both models with various prompts and summary lengths, and opted for \texttt{gpt-3.5-turbo}. While the results did not show significant disparities, \texttt{gpt-3.5-turbo} offered the same quality at just 1/10th of the cost of \texttt{text-DaVinci-003}.\footnote{https://platform.openai.com/docs/models/gpt-3-5}

The temperature was set to zero because we did not need creativity for summarization. 
We employed \texttt{GPT-3.5-turbo} to summarize articles containing fewer than 4,097 tokens, and for articles exceeding 4,097 tokens, we used \texttt{gpt-3.5-turbo-16k}.
See Appendix \ref{appendix:Summarization} for details.

\section{Evaluation}

To assess veracity prediction, we use only automatic metrics including accuracy, precision, recall, and F1 (macro and weighted). 
To assess explanations, we use both automatic and human evaluation methods, in keeping with the recommendation of \citet{Local-Interpretations} that NLE should include human evaluation alongside automatic evaluation.
Note that gold explanations often exhibit a more abstractive nature than explanations generated by LLMs, even when employing abstractive methods for explanation generation. By employing human evaluation, we try to overcome the difficulty of automatically comparing abstractive explanations.

\subsection{Automatic Evaluation}\label{section:autoEval}
For evaluating explanation generation, the correlation between human and automatic 
metrics 
is generally quite low \cite{Explainable-Healthcare-QA}. Nevertheless, following almost all recent related work, we still compare the generated explanation to the gold explanation using ROUGE \cite{lin-2004-rouge}.
 
ROUGE is problematic for comparing abstractive explanations because it is based on exact matching. Natural Language Inference (NLI) has emerged as an alternative method \cite{NLI-Measure-for-Abstractive-Summarization}. 
One advantage of this approach is that it eliminates the need for gold standard explanations. Following \citet{ExClaim} and \citet{Public-Health}, we make use of NLI models to implement reference-free metrics for evaluating the generated text from our NLE models. \citet{Public-Health} introduce the following three NLI-based metrics: 

\paragraph{ \textbf{S}trong \textbf{G}lobal \textbf{C}oherence (SGC).} Every sentence in the 
explanation must entail the claim.

\paragraph{ \textbf{W}eak \textbf{G}lobal \textbf{C}oherence (WGC).} All sentences in the 
explanation 
should either entail or maintain a neutral relation to  the claim. Thus, no sentence in the 
explanation 
should contradict the claim.\footnote{In line with \citet{Public-Health}, for claims originally labeled as \textit{false}, the NLI labels are considered \textit{neutral} if their explanations \textit{contradict} the claim, e.g. we consider the NLI label to be neutral for the following sentence with regard to the claim which was labeled as false originally:

\noindent \textbf{Claim:} \textit{Four 
kids 
who took the 
coronavirus 
vaccine died immediately.}
\textbf{Explanation sentence:} \textit{The claim that four 
children 
died immediately after taking the coronavirus vaccine is false.}}

\paragraph {\textbf{L}ocal \textbf{C}oherence (LC).} In an explanation, no two sentences should contradict each other.

Unfortunately, the implementation of these metrics has not been published, and so we attempt to reproduce them by considering the information provided in \cite{Public-Health}. For each metric, we report the percentage of instances that satisfy the specified metric.

\subsection{Human Evaluation}
To design our human evaluation guidelines, we conducted three iterations of annotation and discussion involving the same two annotators. The final version of the guidelines surpasses the initial one in detail and includes illustrative examples to clarify expectations, leading to an improvement in the inter-annotator agreement.
Guided by these pilot studies, a team of five annotators\footnote{All five were fluent English speakers, with two native speakers and three with English as a second language.} used the guidelines to evaluate explanations, focusing on the following seven criteria:

\paragraph{ \textbf{Repetition of Claim}.} \textit{Is the claim text repeated in the generated explanation?}
This (yes/no) criterion captures the extent to which LLMs repeat the language of the claim in the explanation.

\paragraph{ \textbf{Internal Repetition}.} \textit{Does the generated explanation contain repeated information?}
This yes/no criterion captures one of the common problems with text generation models -- repetition.

\paragraph{ \textbf{Suggested Class}.} \textit{According to the generated explanation, how would you classify the claim, using \textit{true}, \textit{false}, \textit{mixture} and \textit{unproven} labels?}
A generated explanation can be deemed of good quality if, after reading the explanation, the annotator can accurately predict the veracity of the claim. 

\paragraph{ \textbf{Internal Consistency}.} \textit{Is the generated explanation internally consistent, i.e. consistent with itself? An explanation should be considered internally consistent if it does not include a contradiction (includes two statements that contradict each other).} A Likert scale ranging from 0 to 4, where higher scores indicate better quality, was employed.

\paragraph{ \textbf{External Consistency}.}  \textit{Is the generated explanation externally consistent, i.e. consistent with the context? An explanation should be considered externally consistent if it does not include a statement(s) that contradicts a statement(s) in the context.}
As with the Internal Consistency criterion, a Likert scale from 0 to 4 was used.

\paragraph{\textbf{Extra Information}.} \textit{Does the generated explanation contain extra information that is not mentioned in the claim or in the context?}
Given the potential of training data leakage when working with LLMs, particularly in in-context learning experiments, we introduce this yes/no criterion to examine the existence of this property in generated explanations.

\paragraph{ \textbf{Missing Information}.} \textit{Is the generated explanation missing information from the context that is important in explaining the veracity of the claim?}
This criterion allows us to verify whether the generated explanation is sufficient or if additional explanation is required. A three-point scale is used.

In our study, Claim Repetition, Internal Repetition, Extra Information and Missing Information are considered to be undesirable properties. Fig.~\ref{annotation-tool-figure} in Appendix~\ref{appendix:Results} shows a screenshot of the annotation tool we developed.\footnote{As the majority of SOTA LLMs demonstrate high fluency, and based on our pilot studies, we chose to exclude fluency as one of our evaluation criteria.}

\section{Results}

We present the results of automatic as well as human evaluations. Examples of model explanations are provided in Table~\ref{generated-examples} in Appendix~\ref{appendix_example_explanations}.

\subsection{Automatic Evaluation}
The veracity prediction F1 scores for the single and joint tasks are shown in Table \ref{veracity-results}.\footnote{See Table \ref{veracity-detailed-results} in Appendix~\ref{appendix:Results} for precision/recall/accuracy.} For the few-shot setting, we individually selected the best shot number for each model and task on the validation set. In the zero-shot setting, the closed-source models clearly outperform the open-source models, whereas the difference is smaller in the few-shot setting. Fine-tuning achieves the best outcome, particularly fine-tuning of the \textit{Mistral-7B} model, which achieves a macro-F1 of 72.0, slightly higher than the veracity prediction macro-F1 of 70.52 reported by \citet{Public-Health} on the same dataset.
In both zero-shot and few-shot scenarios, the macro F1 for the joint task generally surpasses that of the veracity task, except for the zero-shot performance of \textit{GPT-3.5-D}, the few-shot performance of \textit{Falcon-180B} and zero-shot and few-shot instances of \textit{Llama-70B}. In these cases, the veracity prediction task achieves a higher macro F1 compared to the joint task.

\begin{table}[ht!]
    \centering
    \resizebox{0.48\textwidth}{!}
    {
    \begin{tabular}{clcc}
        \toprule
        \multirow{2}{*}{{\textbf{\rotatebox[origin=c]{90}{\small Setting}}}} &
        \textbf{Task} &
        \textbf{Veracity Pred.} &
        \textbf{Joint Task}\\
        \cmidrule(lr){2-4}
        & \textbf{Model} &
        \textbf{M-F1 / W-F1} & \textbf{M-F1 / W-F1}\\ 
         \midrule
         \multirow{7}{*}{{\textbf{\rotatebox[origin=c]{90}{\small Zero-shot}}}}
          & GPT-3.5-D & 51.7 / 67.8 & 50.0 / 65.9  \\
          & GPT-3.5-T & 51.4 / 69.3 & \textbf{53.9} / \textbf{70.7}\\
          
          & GPT-4 & \textbf{53.2} / \textbf{69.8} & 53.4 / 69.6 \\          
          \cmidrule(lr){2-4}        
          & Falcon-180B & \textbf{36.6} / \textbf{59.0} & 44.2 / \textbf{66.6}\\
          
          & Llama-70B & 33.8 / 49.4 & 31.2 / 46.2 \\
                
          & Vicuna-13B & 23.2 / 24.5 & \textbf{47.4} / 61.4 \\
          
          & Mistral-7B & 20.5 / 25.0 & 41.5 / 55.5 \\          
         \midrule
         
         \multirow{7}{*}{\rotatebox[origin=c]{90}{\textbf{\small Few-shot}}}
          & GPT-3.5-D [4/1] & 49.9 / 67.7 & \textbf{56.6} / \textbf{72.9} \\

          & GPT-3.5-T [2/7] & 52.9 / \textbf{70.1} & 54.5 / 67.5 \\

          & GPT-4 [2/9] & \textbf{53.0} / 69.7 & 54.9 / 71.5 \\

          \cmidrule(lr){2-4}     
          & Falcon-180B [2/1] & \textbf{57.9} / \textbf{74.8} & 51.2 / 70.0 \\

          & Llama-70B [4/4] & 49.3 / 68.6 & 49.0 / 72.6 \\

          & Vicuna-13B [6/7] & 52.4 / 69.7 & \textbf{54.8} / \textbf{75.0}\\

          & Mistral-7B [9/6] & 44.9 / 67.9 & 51.6 / 81.8 \\
         \midrule
         \multirow{2}{*}{\rotatebox[origin=c]{90}{\textbf{\small PEFT}}}
      
          & Vicuna-13B & 68.5 / 80.5 & 70.0 / 81.2 \\
          
          & Mistral-7B & \textbf{72.0} / \textbf{82.5} & \textbf{70.1} / \textbf{82.0}\\
        \bottomrule
    \end{tabular}
    }
  \caption{Test set performance in the veracity prediction and joint tasks, in terms of macro F1 (M-F1) and weighted F1 (W-F1). The designated shot number for each model is specified next to the model name, with the first corresponding to the veracity prediction task and the second to the joint task. }
  \label{veracity-results} 
\end{table}

The Rouge scores for both the single and joint models for explanation generation are reported in Table \ref{ROUGE-results}. Overall, we can observe that the Rouge scores are higher in the few-shot settings compared to the zero-shot setting for both tasks. The highest scores are obtained using PEFT.

\begin{table}[ht!]
    \centering
    \resizebox{0.48\textwidth}{!}
    {
    \begin{tabular}{clcc}
        \toprule
        \multirow{2}{*}{{\textbf{\rotatebox[origin=c]{90}{\small Setting}}}} &
        \textbf{Task} &
        \textbf{Exp. Task} &
        \textbf{Joint Task}\\
        \cmidrule(lr){2-4}
        & \textbf{Model} &
        \textbf{R1 / R2 / RL} & \textbf{R1/ R2 / RL}\\ 
         \midrule
         \multirow{7}{*}{{\textbf{\rotatebox[origin=c]{90}{\small Zero-shot}}}}
          & GPT-3.5-D & 25 / 07 / 16 & 26 / 08 / 17 \\
          & GPT-3.5-T & 28 / 09 / 18 & 26 / 08 / 17 \\
          & GPT-4 & 25 / 07 / 16 & 26 / 08 / 17 \\
          \cmidrule(lr){3-4}
          & Falcon-180B & 22 / 07 / 14 & 18 / 05 / 13 \\
          & Llama-70B & 19 / 06 / 13 & 23 / 07 / 16 \\
          & Vicuna-13B & 22 / 07 / 14 & 24 / 08 / 16\\
          & Mistral-7B & 20 / 06 / 12 & 23 / 07 / 15\\ 
         \midrule
         
         \multirow{7}{*}
         {{\textbf{\rotatebox[origin=c]{90}{\small Few-shot}}}}
          & GPT-3.5-D [1/1] & 25 / 07 / 16 & 24 / 07 / 17 \\
          & GPT-3.5-T [5/7] & 25 / 08 / 16 & 27 / 09 / 19 \\
          & GPT-4 [11/9] & 26 / 09 / 18 & 27 / 09 / 18 \\
          \cmidrule(lr){3-4}
          
          & Falcon-180B [1/1] & 19 / 05 / 12 & 19 / 05 / 12 \\
          & Llama-70B [4/4] & 24 / 09 / 18 & 24 / 08 / 17 \\
          & Vicuna-13B [5/7] & 23 / 07 / 14 & 26 / 08 / 17\\
          & Mistral-7B [3/6] & 23 / 07 / 16 & 24 / 08 / 16\\          
         \midrule
         \multirow{2}{*}{{\textbf{\rotatebox[origin=c]{90}{\small PEFT}}}}        
          & Vicuna-13B & \textbf{
          36 / 15 / 27} & \textbf{36 / 15 / 27}\\
          & Mistral-7B & 34 / 14 / 25 & 36 / 15 / 26\\
        \bottomrule
    \end{tabular}
    }
  \caption{ROUGE-1 (R1), ROUGE-2 (R2), ROUGE-L (RL) F1 scores on the test set for generated explanations}
  \label{ROUGE-results} 
\end{table}

\begin{table*}[ht!]
    \setlength{\tabcolsep}{10pt}
    \resizebox{\textwidth}{!}
    {
    \begin{tabular}{clrrrrrrrr}
        \toprule
        \multirow{2}{*}{\textbf{\rotatebox[origin=c]{90}{\small Setting}}} &
        \multirow{2}{*}{\textbf{Model}} &
        \multicolumn{4}{c}{\textbf{Exp. Task }} &
        \multicolumn{4}{c}{\textbf{Joint Task}}\\
        \cmidrule(lr){3-6}
        \cmidrule(lr){7-10}
         & &
        \textbf{SGC} & \textbf{WGC} & \textbf{LC} & \textbf{S-Score} &
        \textbf{SGC} & \textbf{WGC} & \textbf{LC} & \textbf{S-Score}\\
         \midrule
         & Gold Explanations & - & -& - & - & 22.0 & 93.02 & 75.24 & -\\
         \midrule  
         \multirow{7}{*}{\textbf{\rotatebox[origin=c]{90}{\small Zero-shot}}}
          & GPT-3.5-D & 3.73 & 90.18 & 87.66 & 53.09 & 12.66 & 90.67 & 90.91 & 51.92 \\
         
          & GPT-3.5-T & 0.24 & 84.21 & 42.11 & 51.11 & 00.97 & 88.88 & 81.90 & 53.42 \\
         
          & GPT-4 & 1.41 & 92.74 & 81.03 & \textbf{54.37} & 5.19 & 90.58 & 87.50 & \textbf{53.60} \\
         \cmidrule(lr){2-10}

        & Falcon-180B & 4.38 & 81.09 & 57.95 & 47.55 & 2.52 & 91.23 & 78.49 & 48.16\\
         
          & Llama-70B & 4.00 & 83.75 & 61.77 & \textbf{48.38} & 4.06 & 82.73 & 76.29 & 40.28\\
         
          & Vicuna-13B & 0.00 & 78.49 & 44.89 & 46.25 & 0.81 & 86.61 & 66.8 & \textbf{49.35} \\
         
          & Mistral-7B & 0.24 & 75.00 & 33.93 & 43.50 & 0.89 & 81.74 & 55.44 & 44.94 \\
         \midrule
         
         \multirow{7}{*}{\textbf{\rotatebox[origin=c]{90}{\small Few-shot}}}
          & GPT-3.5-D [1/1] & 2.19 & 89.29 & 84.42 & 52.65 & 28.41 & 93.75 & 98.62 & \textbf{55.99} \\
         
          & GPT-3.5-T [5/7] & 2.52 & 90.02 & 79.87 & 53.01 & 15.99 & 91.40 & 93.99 & 54.85\\
         
          & GPT-4 [11/9] & 15.26 & 90.58 & 82.63 & \textbf{54.29} & 13.64 & 91.31 & 89.04 & 54.78\\
         \cmidrule(lr){2-10}
         
         & Falcon-180B [1/1] & 0.00 & 80.60 & 43.02 & 46.80 & 0.08 & 82.39 & 44.81 & 48.20 \\
         
        & Llama-70B [4/4] & 30.35 & 94.27 & 81.11 & \textbf{56.14} & 20.54 & 88.64 & 64.37 & 48.41\\
         
          &Vicuna-13B [5/7] & 0.97 & 78.73 & 43.51 & 46.37 & 7.39 & 85.96 & 70.86 & \textbf{50.64}\\
         
          & Mistral-7B [3/6] & 8.85 & 87.18 & 71.27 & 51.59 & 7.06 & 83.77 & 45.37 & 48.24\\
         \midrule
         
         \multirow{2}{*}{\textbf{\rotatebox[origin=c]{90}{\small PEFT}}}
          & Vicuna-13B & 30.52 & 93.99 & 75.57 & \textbf{60.50} & 25.00 & 92.69 & 73.54 & \textbf{64.94}\\
         
          & Mistral-7B & 23.13 & 93.18 & 75.89 & 59.09 & 26.70 & 92.21 & 76.70 & 64.61 \\
          \bottomrule         
    \end{tabular}
   }
  \caption{NLI-based coherence metrics on the test set for explanation generation and the joint task using the \cite{forth-nli-method} NLI model.}    
  \label{coherence-results}
\end{table*}

The NLI-based coherence metrics described in Section~\ref{section:autoEval} are calculated using four different NLI models, 1) a decomposable attention model \cite{DAELMO}, 2) a RoBERTa model trained SNLI \cite{RoBERTa-snli}, 3) a RoBERTa model trained on MNLI \cite{RoBERTa-mnli}, and 4) a SOTA NLI model pretrained on various NLI datasets using RoBERTa \cite{forth-nli-method}. The first three models were used by \citet{Public-Health}.\footnote{Since \citet{Public-Health} have not released either their implementation of the coherence metrics or the generated results on the test set, a comparison is difficult.}
Table \ref{coherence-results} shows the coherence metrics results on the test using only the \cite{forth-nli-method} NLI model.
The detailed results obtained using all four NLI models are presented in Table \ref{coherence-full-results} in Appendix~\ref{appendix:Results}. As with the Rouge score, the majority of models benefit from moving from zero-shot to few-shot, particularly \textit{Llama-70B}. An exception is \textit{Falcon-180B}.

In order to choose a subset of the models for our human evaluation study, we categorize all LLMs into five groups: zero-shot closed- and open-source, few-shot closed- and open-source, and PEFT. 
We also define a new metric for choosing the best model in each category automatically, the \textit{Selection Score} or \textbf{S-Score} in Table~\ref{coherence-results}. In the explanation task, the S-Score is computed as the highest mean of RougeL F1 and WGC; in the joint task, it is computed as the weighted mean of the macro F1 (veracity prediction), WGC, and RougeL F1, with the respective weights of 0.5, 0.25, and 0.25.

\subsection{Human Evaluation}
\label{human-evaluation-result-section}
\begin{table*}[b!p]
\small
\begin{tabular}{p{15.5cm}}
\toprule
\multicolumn{1}{c}{\textbf{Example 1}} \\
\midrule
\textbf{Claim:} The Obama administration shut down the Amber Alert program because of the government shutdown.

\textbf{Summarized Context:} 
The claim that the Obama administration shut down the Amber Alert program due to the government shutdown is false. The Amber Alert system is not run by the federal government, but rather by individual states and coordinated through the National Center for Missing and Exploited Children, a private nonprofit organization. The federal webpage promoting the program was temporarily closed due to the government shutdown, but this did not affect the operation of the program itself. The Amber Alert system continued to operate during the shutdown, with alerts being sent out in Texas and Florida. The federal webpage was restored to eliminate confusion among the public, but it was purely informational and did not impact the functioning of the program. The Amber Alert system is activated by local or state law enforcement officials when they have specific details about a missing child, such as their name, photo, and information about the suspect or vehicle involved. The system relies on media and cell phone notifications to spread the word and has been successful in locating missing children in the past. The federal webpage has never been involved in the actual process of finding missing children. Overall, the claim that the Obama administration shut down the Amber Alert program is false and based on a misunderstanding of how the system operates.
        \\
        \midrule
        \multicolumn{1}{c}{\textbf{GPT4 (zero-shot) Explanation Task}} \\
        \midrule
        The claim is false because the Amber Alert system was not shut down during the government shutdown. The system is run by individual states and coordinated through a private nonprofit organization, not the federal government. The shutdown only affected the federal webpage promoting the program, not the actual operation of the Amber Alert program. Alerts were still sent out in different states during the shutdown. The federal webpage was purely informational and does not impact the functioning of the program. Thus, the claim that \textcolor{cyan}{the Obama administration shut down the Amber Alert program because of the government shutdown} is not accurate.\\
        \midrule
        \multicolumn{1}{c}{\textbf{Llama-70B ( Zero-shot) Explanation Task}} \\
        \midrule
        The claim that \textcolor{cyan}{the Obama administration shut down the Amber Alert program is false} because the Amber Alert system is not run by the federal government, but rather by individual states and coordinated through a private nonprofit organization. The federal webpage promoting the program was temporarily closed due to the government shutdown, but this did not affect the operation of the program itself. The Amber Alert system continued to operate during the shutdown, with alerts being sent out in Texas and Florida. The federal webpage was restored to eliminate confusion among the public, but it was purely informational and did not impact the functioning of the program. The Amber Alert system is activated by local or state law enforcement officials when they have specific details about a missing child, such as their name, photo, and information about the suspect or vehicle involved. The system relies on media and cell phone notifications to spread the word and has been successful in locating missing children in the past. The federal webpage has never been involved in the actual process of finding missing children. \textcolor{cyan}{Overall, the claim that the Obama administration shut down the Amber Alert program is false} and based on a misunderstanding of how the system operates. \\
        \midrule
        \multicolumn{1}{c}{\textbf{GPT4 (few-shot) Explanation Task}} \\
        \midrule
        The US federal shutdown did not shut down the amber alert system, despite incorrect reports to the contrary. The official US government amber alert website, which promotes the program and educates the public about amber alerts, was taken offline, but this did not affect the functioning of the amber alert system itself. Amber alerts are actually organized, activated, and issued at the state and local level, and they continued to be issued during the shutdown. Amber Alerts are emergency messages that are broadcast when a law enforcement agency determines that a child has been abducted and is in imminent danger. They are broadcast via radio, television, road signs, cell phones, and other data-enabled devices. While it’s important to note that the federal website is not the amber alert system. The federal website promotes the program and issues guidelines, but it has no active role in issuing amber alerts. While it’s also correct to observe that funding for the amber alert program comes from a mixture of federal, state, and local resources. The office of justice programs, which oversees the amber alert program, is not paid directly by the federal government. Instead, its funding flows from the Department of Justice to state and local partners. Therefore, even during a federal shutdown, essential public safety programs such as the amber alert program would continue to operate.\\
        \midrule
        \multicolumn{1}{c}{\textbf{Llama-70B (few-shot) Explanation Task}} \\
         \midrule
        The Amber Alert program was not shut down during the government shutdown.\\
        \bottomrule
        \end{tabular}
        \caption{GPT-4 and Llama explanations (explanation task only) See Table~\ref{generated-examples} in Appendix~\ref{appendix:Results} for more model outputs for this example, and for more examples. \textcolor{cyan}{Claim repetition, internal repetition, or copy context as explanation}}\label{short-generated-examples}
        \end{table*}
Given that human evaluation of generated texts is an arduous and costly process, we limit our evaluations to the best LLM from each category according to the S-Score in Table \ref{coherence-results}.
That results in ten settings for the five model categories across the two tasks.
The evaluation set consists of 52 instances from the test set for each model, sampled so as  to follow the distribution of classes: 31 instances of \textit{True}, 14 of \textit{False}, 4 of \textit{Mixture}, and 3 of \textit{Unproven}. 
In total, we assess 520 instances, with an additional 10\% overlap for agreement calculation (resulting in 572 instances). 
The manual evaluation process required around 250 hours of annotation work. 
The findings reveal robust inter-annotator agreement, particularly for \textit{Internal Repetition} and \textit{Extra Information}, where we demonstrated over 94\% concordance. 
Agreement rates exceeded 82\% across all other criteria, with the exception of \textit{Missing Information}, which still exhibited a respectable 71\% agreement (See Table \ref{human_eval-agreement_table} in the Appendix).

To more easily compare models for each task according to our human evaluation protocol, we introduce three new scores: S3, S5, and S7. 
The S3 score represents the percentage of instances satisfying the Extra Information, Missing Information, and Suggested Class criteria. 
The S5 score indicates the percentage of instances meeting both Internal Consistency and External Consistency criteria in addition to those in S3. 
The S7 score is the most comprehensive score, indicating the percentage of instances that fulfil all seven criteria. 

Table~\ref{human-evaluation-results} shows these scores for each of the ten selected models. 
According to the S7 metric, the few-shot \textit{GPT-4} model emerges as the optimal choice overall for generating high-quality explanations in the explanation task. 
For the joint task, the few-shot \textit{GPT-3.5-D} and \textit{Vicuna-13B} models show promising performance. 
Comparing the number of parameters in 
\textit{Vicuna-13B} with those in \textit{GPT-3.5-D}, \textit{GPT4}, or 
\textit{Llama-70B}, the performance of the \textit{Vicuna-13B} model after 
parameter-efficient fine-tuning (PEFT) is noteworthy.

We also present the results for the gold standard 
explanation (last row in Table \ref{human-evaluation-results}). 
Interestingly, most of the LLMs perform notably better than the gold standard, suggesting that human generated abstractive explanations are not always of good quality. 
Considering the low scores of the gold explanations, it is perhaps 
unsurprising that few-shot scenarios outperform PEFT in generating explanations for both tasks.

There are a few possible reasons for the lower scores of gold explanations from the PUBHEALTH dataset.
Firstly, all claims and related context, veracity labels and explanations were collected automatically from different fact-checking websites and the PUBHEALTH  authors mapped almost 100 labels into four labels (True, False, Mixture, and Unproven). 
While fact-checking websites generally share a common definition for fact-checking labels, there can still be slight differences. 
Mapping all labels into four categories could certainly introduce ambiguity. 
The gold explanations exhibit a notable MAE of 50\% for the Extra Information criterion, whereas the worst performing models for the same criterion introduce extra information in 38\% of their generated explanations. 
This suggests that in real-world scenarios, journalists assess each claim using multiple, diverse sources, sometimes relying on their own knowledge, rather than solely relying on the provided article or context. 
Moreover, automatic instance crawling can inevitably introduce some noise into the dataset.  

Tables~\ref{human-evaluation-detailed-results} and \ref{detailed-human-evaluation-results} (see Appendix) report the error rate per criterion, model-wise breakdown and results per class. The correlations between automated metrics and human-evaluated explanation quality remain consistently weak, corroborating previous findings ~\cite{Local-Interpretations} -- see \autoref{correlation-result-table}.

\begin{table}[ht!]
    \centering
    \resizebox{0.48\textwidth}{!}
    {
    \begin{tabular}{cllccc}
        \toprule
        &
        \textbf{Setting} &
        \textbf{Model} &
        \textbf{S3} &
        \textbf{S5} &
        \textbf{S7}\\ 
        \midrule
         \multirow{5}{*}{\rotatebox[origin=c]{90}{\textbf{Explanation}}} & \multirow{2}{*}{Zero-shot} &
          GPT-4 & \textbf{76.92} & \textbf{73.08} & 36.54 \\
           & & Llama-70B & 65.38 & 65.38 & 23.08 \\
          \cmidrule(lr){2-6}  
        
          & \multirow{2}{*}{Few-shot} &
          
          GPT-4 & 42.31 & 42.31 & \textbf{38.46} \\
 
            & & Llama-70B & 32.69 & 32.69 & 25.00 \\
         \cmidrule(lr){2-6}

          & PEFT &
          
          Vicuna-13B & 36.54 & 36.54 & 25.00 \\
         \midrule
         
         \multirow{6}{*}{\rotatebox[origin=c]{90}{\textbf{Joint}}} & \multirow{2}{*}{Zero-shot} &
         
           GPT-4 & 59.62 & 57.69 & 38.46 \\
           
           & & Vicuna-13B & 55.77 & 51.92 & 25.00 \\
          \cmidrule(lr){2-6}  
          
          & \multirow{2}{*}{Few-shot} &
          
          GPT-3.5-D & 51.92 & 51.92 & \textbf{48.08} \\

           & & Vicuna-13B & \textbf{67.31} & \textbf{67.31} & \textbf{48.08} \\
         \cmidrule(lr){2-6}

          & PEFT &
          Vicuna-13B & 42.31 & 42.31 & 40.38 \\
          \cmidrule(lr){2-6}

          & Gold Exp. & & 25.00 & 25.00 & 19.23\\
        \bottomrule
    \end{tabular}
    }
  \caption{Human evaluation results: S3 denotes the percentage of instances meeting Extra Information, Missing Information, and Suggested Class criteria; S5 indicates the percentage of instances fulfilling Internal Consistency and External Consistency criteria in addition to those in S3; and S7 represents the percentage of instances meeting all seven criteria.}
  \label{human-evaluation-results} 
\end{table}

\paragraph{Analysis.}
After conducting a comprehensive manual inspection of over 250 
explanations, along with their manually annotated scores according to our evaluation criteria, we conclude that no single criterion can definitively determine the superiority of one model over another. For instance, the zero-shot \textit{GPT-4} model performs exceptionally well in generating high-quality explanations, if the Suggested Class criterion is considered (F1 score of 78.79). However, it is noteworthy that models may achieve high accuracy by merely repeating the claim without offering substantial explanations (see the generated explanation of few-shot \textit{Llama-70B} for the explanation task in Table \ref{short-generated-examples} as an example). Additionally, the zero-shot \textit{GPT-4} model also exhibits the highest Claim Repetition and Internal Repetition scores at 44.23\% and 17.31\%. We therefore advocate for a comprehensive approach which considers all criteria simultaneously (S7), while also reporting S3 and S5 as more lenient metrics.

In the explanation task, models attempt to summarize the context as the explanation. We believe this behavior stems from the task's nature, where we provide the veracity label of the claim along with the claim and context, and inquire about the reasons behind the veracity label. In contrast, in the joint task, we solely input the claim and context, prompting the models to predict the veracity label and provide reasons for their prediction. Consequently, the explanations generated in the joint task exhibit higher realism and quality compared to those in the explanation task. Models appear to seek relevant information from the context to generate the rationale behind their predictions. This explains the improvement observed across all criteria, except for Suggested Class\footnote{The lack of improvement here is reasonable as we do not provide the gold veracity label as input.}  when comparing the results of models in the joint task to their counterparts in the explanation task. Furthermore, in the joint task, models generally produce shorter yet more accurate explanations compared to the explanation task. This observation is consistent with the average number of generated words across all models and test set instances --
94 words for the joint task and 123 words for the explanation task.

According to the relaxed scores (S3 and S5) in Table \ref{human-evaluation-results}, zero-shot models outperform few-shot models, especially in the explanation task. For instance, in the explanation task, the zero-shot scenario of \textit{Llama-70B} performs better than its few-shot counterpart. This discrepancy arises because the relaxed scores overlook the Claim Repetition and Internal Repetition criteria. In the zero-shot scenario, especially for open-source LLMs, some instances involve the model simply duplicating the context or claim without providing meaningful explanations, or just 
regenerating/predicting the veracity label of the claim beside the claim without any explanation. Consequently, the 
relaxed scores of these models in the zero-shot scenario are higher than in the few-shot scenario, because Claim Repetition and Internal Repetition do not contribute to the scores. However, when considering the perfect score (S7), we observe the opposite trend, with few-shot outperforming zero-shot.

Another noteworthy observation is that some models encounter difficulties in providing explanations for instances with Unproven claim veracity labels, generating unrelated text that is relevant neither to the claim nor the context (see the third example in Table \ref{generated-examples} in Appendix~\ref{appendix:Results}). Furthermore, after reviewing the confusion matrix for each model (see Table \ref{confusion_matrix_table} in Appendix~\ref{appendix:Results}), we observe instances where models misclassify the True, False, and Mixture classes as Unproven. This occurs when models either introduce 
information not present in the context or overlook crucial information in the context (Figure \ref{merged_heatmap} illustrates the heatmap depicting the correlation between various evaluation criteria).

\section{Conclusions}
We have presented a set of novel explainable fact-checking experiments 
with closed- and open-source LLMs in a variety of settings, offering valuable insights into LLMs' performance in claim verification and explanation within the public health domain, 
A second contribution of this paper is the human evaluation of the generated explanations and a novel set of evaluation guidelines. As well as highlighting differences between the models, the human evaluation reveals some issues with the gold explanations in the PUBHEALTH dataset.

\section{Limitations}
We note the following limitations:
\begin{enumerate}
\setlength\itemsep{-0.5em}
\item Fine-tuning of Llama-70B and Falcon-180B was not possible due to computational budget limitations. This means that our fine-tuning was restricted to the Mistral-7B and Vicuna-13B models.
\item Our experiments were focused on the English language and the public health domain.

\item We have conducted a human evaluation with five annotators, 10 models, and 52 samples for each model, totaling 520 instances manually inspected. This required much effort (around 250 hours) but there is always room for more qualitative analysis. 
\end{enumerate}
\section*{Acknowledgements}
This research is supported by Science Foundation Ireland (SFI)
through the SFI Frontiers for the Future programme (19/FFP/6942) and the ADAPT Centre for Digital Content Technology, which is
funded under the SFI Research Centres Programme (Grant 13/RC/2106)
and is co-funded under the European Regional Development.
We thank the reviewers for their insightful and helpful comments.

\bibliography{tacl2021}
\bibliographystyle{acl_natbib}

\appendix
\section{Experimental Details}\label{appendix:experiment-details}
\subsection{Zero-shot and Few-shot Details}
\label{appendix:Zero-Few-shots}

\subsubsection{Prompts}
We used various prompts for each task including veracity, explanation, and joint. In addition, different prompts were utilized for closed-source and opened-source LLMs. As a result, we employed a range of prompts for each task on a small subset and manually assessed the results. Subsequently, we selected the most promising prompt and further refined it using the https://claude.ai engine to enhance its effectiveness. The final experimented prompts for closed-source LLMs are mentioned in the section \ref{Zero-Few-shots-main-section}, and the final experimented prompts for opened-source LLMs are as follows: 

\textbf{Veracity Prediction:} \texttt{\#\#\# Instruction:\textbackslash nUse the Task below and the Input given to write the Response, which is a veracity label prediction that can solve the Task. \textbackslash n \textbackslash n\#\#\# Task:\textbackslash nBased only on the context, categorize the claim as: \textbackslash nTrue (supported by context) \textbackslash n False (contradicted by context) \textbackslash nMixture (partially supported/contradicted) \textbackslash nUnproven (not enough info) \textbackslash nOnly generate a single word as response. \textbackslash n \textbackslash n\#\#\# Input:\textbackslash nContext: X \textbackslash nClaim: Y \textbackslash n \textbackslash n\#\#\# Response: \textbackslash n}

\textbf{Explanation Generation:} \texttt{\#\#\# Instruction:\textbackslash nUse the Task below and the Input given to write the Response, which is an explanation generation that can solve the Task. \textbackslash n \textbackslash n\#\#\# Task:\textbackslash nUsing only the context provided, explain why the claim veracity is Z.\textbackslash n \textbackslash n\#\#\# Input:\textbackslash nContext: X \textbackslash nClaim: Y \textbackslash n The claim veracity: Z \textbackslash n \textbackslash n\#\#\# Response: \textbackslash n}

\textbf{Joint Task:} \texttt{\#\#\# Instruction:\textbackslash nUse the Task below and the Input given to write the Response, which is a veracity label prediction and the reason explanation for your prediction that can solve the Task. \textbackslash n \textbackslash n\#\#\# Task:\textbackslash nBased only on the context, categorize the claim as: \textbackslash nTrue (supported by context) \textbackslash n False (contradicted by context) \textbackslash nMixture (partially supported/contradicted) \textbackslash nUnproven (not enough info) \textbackslash nAnd explain your reasoning. Provide the response in JSON format with the following keys: veracity, explanation. \textbackslash n \textbackslash n\#\#\# Input:\textbackslash nContext: X \textbackslash nClaim: Y \textbackslash n \textbackslash n\#\#\# Response: \textbackslash n}

In this context, X, Y, and Z represent the contextual content, claim text, and the veracity label of the claim, respectively

\subsubsection{Few-shot Tuning}
To determine the optimal number of shots, we randomly selected a subset of 100 samples from the dev set, considering class frequency. We conducted experiments covering a range of numbers, from 1-shot to 12-shot (excluding cases where the max sequence length of the LLMs was exceeded), for all three tasks using this subset. This process was repeated three times with three subsets for open-source LLMs for considering potential noises and variances. However, to minimize costs for closed-source LLMs, we only performed these experiments with one subset. In the veracity task, we computed the variance and the mean of macro F1 for each shot number based on the results from three rounds. The one with the highest mean and the lowest variance was selected as the best shot number.

In the explanation task, we selected the shot number based on the highest mean of RougeL F1 and WGC, prioritizing those with low variance across three rounds. Finally, in the joint task, we defined a selection score by calculating the mean of macro F1, WGC, and RougeL F1. By using the veracity section of the results, we assigned fifty percent weight to the mean of macro F1 in the selection score. Simultaneously, the other fifty percent weight in the selection score was given to the mean of RougeL F1 and WGC from the explanation section of the results. Then, the shot number with the highest selection score and lowest variance was selected as the best shot.

\subsection{Setting Details}
\label{appendix:settings}
We conducted zero-shot and few-shot experiments with default hyperparameter values for all selected LLMs. Due to resource constraints, we quantized the Falcon-180B model to 8 bits for our in-context learning experiments. For closed-source LLMs, we set the \textit{max new tokens} to 3 for the veracity task and 300 for the explanation and joint tasks. For open-source LLMs, we adjusted the \textit{max new tokens} to 5, 348, and 360 for the veracity, explanation, and joint tasks, respectively.

\begin{table}[ht!]
    \centering
    \setlength{\tabcolsep}{6pt}
    \resizebox{0.48\textwidth}{!}
    {
    \begin{tabular}{llccc}
        \toprule
        \textbf{Task} &
        \textbf{Model} &
        \textbf{epochs} &
        \textbf{lora\_dropout} &
        \textbf{seq. length}
        \\ 
        \midrule
         \multirow{2}{*}{\textbf{Veracity}} & 
          Vicuna-13B & 10 & 0.45 & 830 \\
            
            \cmidrule(lr){2-5}

           & Mistral-7B & 12 & 0.50 & 830 \\
            
            \midrule
            
          \multirow{2}{*}{\textbf{Explanation}} & Vicuna-13B & 10 & 0.50 & 1700 \\
            
            \cmidrule(lr){2-5}

           & Mistral-7B & 10 & 0.50 & 1700 \\
            
            \midrule

         \multirow{2}{*}{\textbf{Joint}} & Vicuna-13B & 15 & 0.55 & 1700 \\
            
            \cmidrule(lr){2-5}

           & Mistral-7B & 15 & 0.55 & 1700 \\
            
        \bottomrule
    \end{tabular}
    }
  \caption{Hyper-parameter settings for each model and task. seq. length refers to the maximum sequence length for models.}
  \label{hyperparameters-table} 
\end{table}

We conducted parameter-efficient fine-tuning using Vicuna-13B and Mistral-7B models utilizing 4-bit quantization. Our fine-tuning process employed the AdamW (paged\_adamw\_32bit) optimizer with a learning rate of 2e-4, and we fine-tuned our models with various hyperparameter values, selecting the optimal values based on performance on the validation set. For QLoRA settings, we determined the best values for \textit{r} and \textit{alpha} to be 16. Additionally, we configured \textit{bias} and \textit{task\_type} as \textit{none} and \textit{CAUSAL\_LM}, respectively, following the default settings of QLoRA. Refer to Table \ref{hyperparameters-table} for a comprehensive overview of other hyperparameter settings for each model and task.

\section{Summarization Details}
\label{appendix:Summarization}

\subsection{Prompts}
\begin{figure}[ht]
\begin{subfigure}{.5\textwidth}
  \centering  \includegraphics[scale=.32]{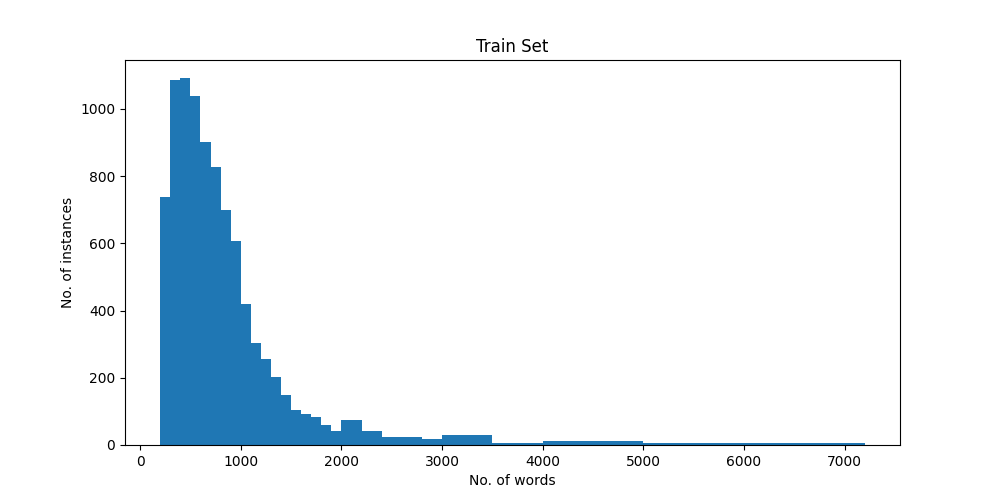}  
  \caption{The number of words in the main text in the train set}
  \label{check-words-no-train}
\end{subfigure}
\begin{subfigure}{.5\textwidth}
  \centering
\includegraphics[scale=.32]{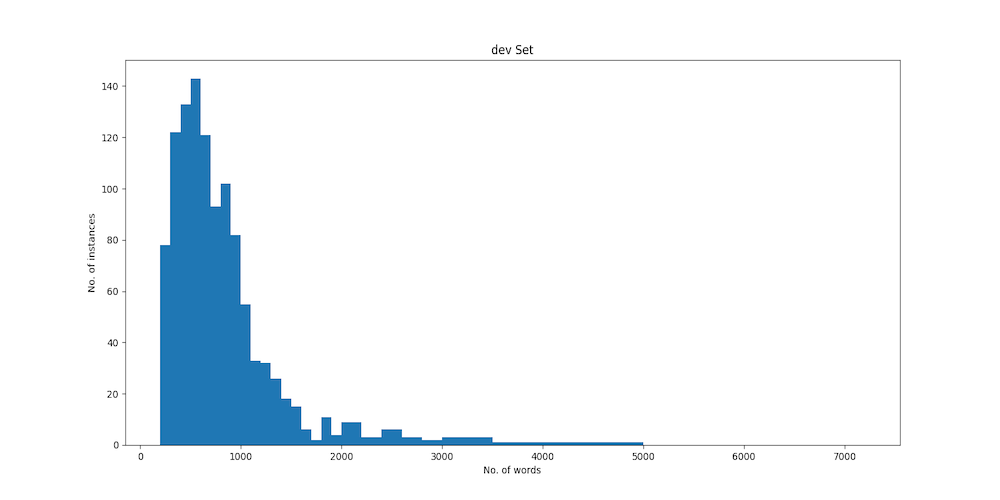}  
  \caption{The number of words in the main text in the validation set}
  \label{check-words-no-dev}
\end{subfigure}
\begin{subfigure}{.5\textwidth}
  \centering
\includegraphics[scale=.32]{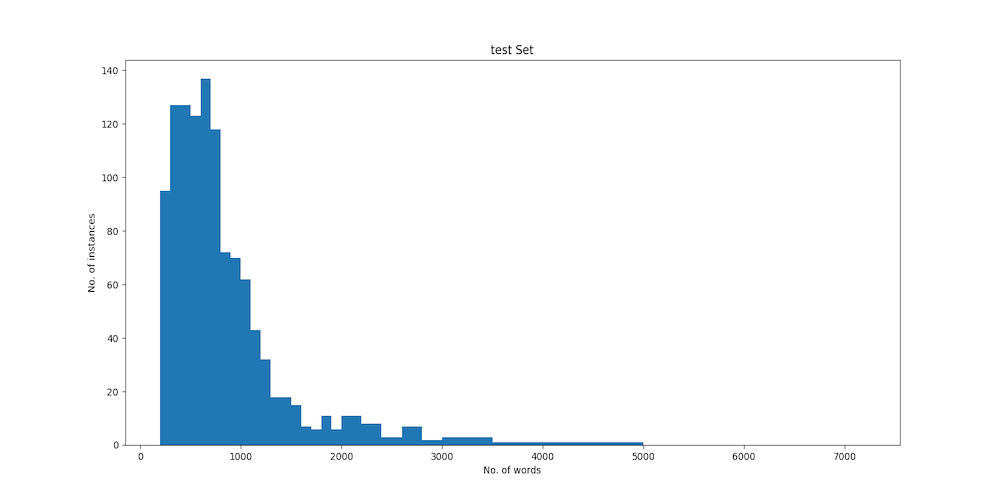}  
  \caption{The number of words in the main text in the test set}
  \label{check-words-no-test}
\end{subfigure}
\caption{The number of words in the main text in different sets of PUBHEALTH dataset}
\label{check-words-no}
\end{figure}

Firstly, we examined the number of words in articles in the PUHEALTH dataset (Figure \ref{check-words-no}). The mean and median word counts across all sets are approximately 700 and 600 words, respectively. Consequently, we tested the length of the summary output with 250 words and 350 words. We randomly selected 14 examples from the PUBHEALTH train set, each featuring varying base word counts, spanning from 600 to 1600 words. After examining this subset manually, we chose to limit the summary output to 350 words. This is because longer summaries contain additional details, ensuring we will not overlook any essential information from the article content regarding the claim for the next steps. Indeed, we utilize the summarized article content and the claim to predict the veracity of the claim and generate an explanation for the veracity prediction. In addition, we did not summarize articles that consist of less than 350 tokens, which resulted in skipping 1,262 samples of the whole PUBHEALTH dataset.

Secondly, we tested various prompts as follows to ask the LLM to summarize the text. We selected prompt number seven after manually comparing the results of all prompts on the selected subset.

\begin{enumerate}
  \item \texttt{Your task is to generate a summary of a news article for use in claim verification.
  Summarize the news article below, focusing on any aspects that are relevant to the claim below. Both claim and news article are delimited by triple backticks. Limit to [250, 350] words.
claim: : ```[]```
news article: ```[]```}
  \item \texttt{Your task is to generate a summary of a news article for use in claim verification.
  Summarize the news article below, focusing on any aspects that are relevant to the claim below. Limit to [250, 350] words.
claim: : ```[]```
news article: ```[]```}
  \item \texttt{Your task is to generate a summary of a news article for use in claim verification.
  Summarize the article below, focusing on any aspects that are relevant to the claim below. Limit to 350 words. Do not assess the veracity of the claim. Do not explain the veracity of the claim.
claim: : ```[]```
news article: ```[]```}
  \item \texttt{Your task is to generate a summary of an article.
Summarize the article below, focusing on any aspects that are relevant to the claim below. Limit to 350 words. Do not assess the veracity of the claim. Do not explain the veracity of the claim.
claim: : ```[]```
news article: ```[]```}
\item \texttt{Your task is to summarize an article.
Extract all important information from the article below, focusing on any aspects that are relevant to the claim below. Limit to 350 words.
claim: : ```[]```
news article: ```[]```}
  \item \texttt{Your task is to extract all important information from an article.
Extract all important information from the article below, focusing on any aspects that are relevant to the claim below. Limit to 350 words. Do not assess the veracity of the claim. Do not explain the veracity of the claim.
claim: : ```[]```
article: ```[]```}

\item \texttt{Your task is to summarize an article.
Extract all important information from the article below, focusing on any aspects that are relevant to the claim below. Limit to 350 words.
claim: : ```[]```
article: ```[]```}
\end{enumerate}

We removed the phrase "\textit{for use in claim verification}" from the prompt because, in our perspective, this phrase could introduce ambiguity to the LLM. Including it might prompt the LLM to assess or explain the claim's veracity rather than concentrating on summarizing the article. After checking the result of the experiments with and without extra rules (\textit{Do not assess the veracity of the claim. Do not explain the veracity of the claim.}), We chose not to implement these rules because, despite the lack of significant differences in the results, the prompt without additional rules was shorter and led to cost savings.

\subsection{Evaluation}
In order to analyze the quality of our summarization process and pick the best model and setting, we evaluated output summaries in the sampled subset manually. Each summary output was evaluated based on
three criteria: coherence, relevance, and missing information. For the first two criteria, we follow \citet{Re-evaluating-Summarization-Evaluation} guidelines. We also consider our definition for the last criterion in the human evaluation of summaries.

\textbf{Coherence:} The summary must demonstrate a clear and organized structure. It should not merely present a collection of related details but instead progress logically from one sentence to another, forming a cohesive body of information extracted from the article text specifically pertaining to the related claim.

\textbf{Relevance:} The summary should encompass solely crucial information extracted from the article text, directly relevant to the claim.

\textbf{Missing information:} Is the generated summary missing essential information from the article text crucial for evaluating or explaining the claim's veracity?

In the end, we chose gpt-3.5-turbo as the summarizer model. This decision was made using the seventh prompt, requesting a 350-word output summary, and setting the temperature to zero.

\begin{figure*}
  \centering
  \includegraphics[scale=.68]{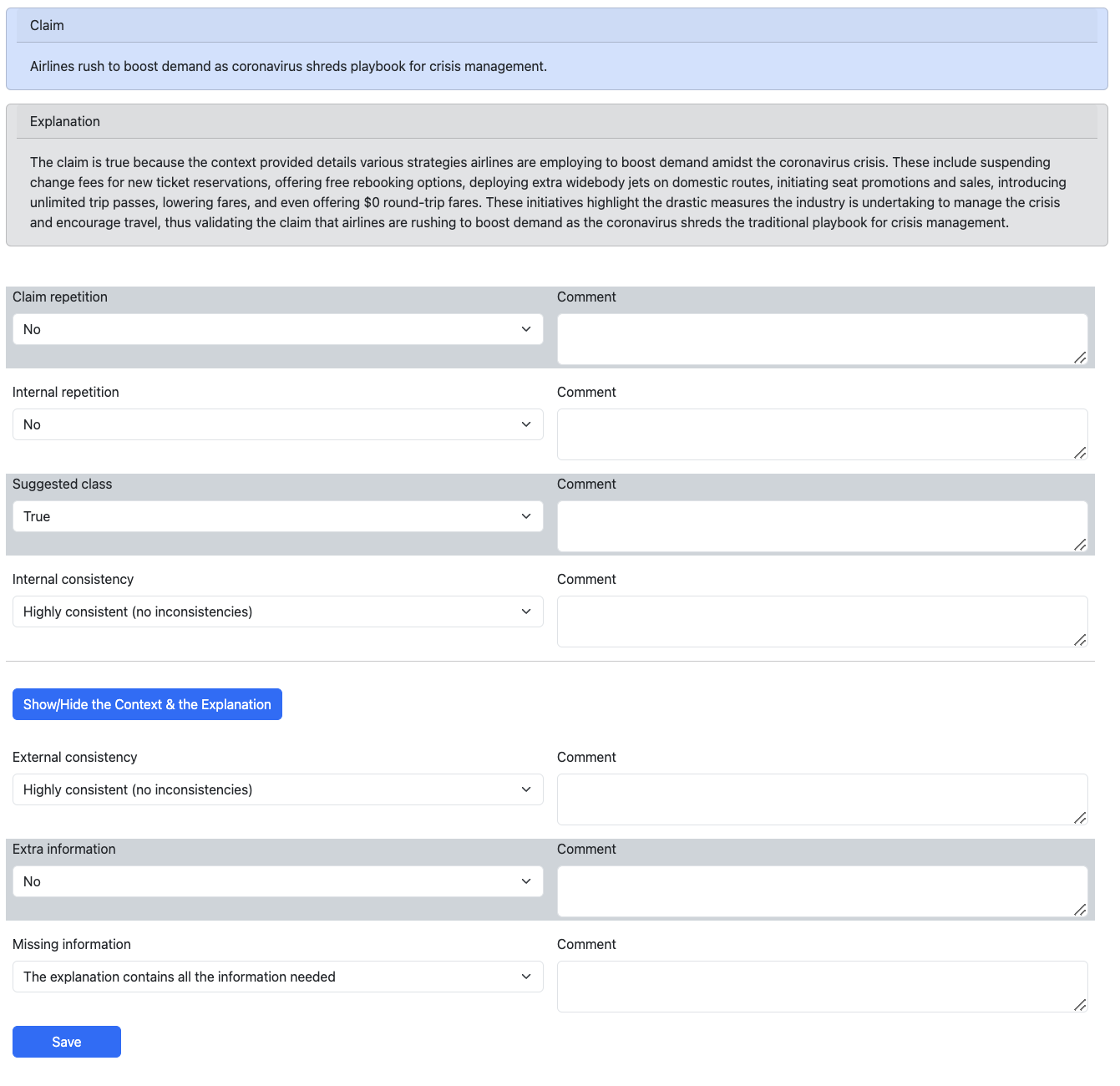}
  \caption{A screenshot of our annotation tool.}
  \label{annotation-tool-figure}
\end{figure*}
\section{Results}
\label{appendix:Results}

\subsection{Examples of Generated Explanations}
\label{appendix_example_explanations}

In Table \ref{generated-examples} we present instances of generated explanations produced by various models.

\begin{table*}[b!p]
\small
\begin{subtable}{\textwidth}
\begin{tabular}{p{15.5cm}}
\toprule
\multicolumn{1}{c}{\textbf{Example 1}} \\
\midrule
\textbf{Claim:} The Obama administration shut down the Amber Alert program because of the government shutdown.

\textbf{Summarized Context:} 
The claim that the Obama administration shut down the Amber Alert program due to the government shutdown is false. The Amber Alert system is not run by the federal government, but rather by individual states and coordinated through the National Center for Missing and Exploited Children, a private nonprofit organization. The federal webpage promoting the program was temporarily closed due to the government shutdown, but this did not affect the operation of the program itself. The Amber Alert system continued to operate during the shutdown, with alerts being sent out in Texas and Florida. The federal webpage was restored to eliminate confusion among the public, but it was purely informational and did not impact the functioning of the program. The Amber Alert system is activated by local or state law enforcement officials when they have specific details about a missing child, such as their name, photo, and information about the suspect or vehicle involved. The system relies on media and cell phone notifications to spread the word and has been successful in locating missing children in the past. The federal webpage has never been involved in the actual process of finding missing children. Overall, the claim that the Obama administration shut down the Amber Alert program is false and based on a misunderstanding of how the system operates.
        \\
        \midrule
        \multicolumn{1}{c}{\textbf{GPT4 (zero-shot) Explanation Task}} \\
        \midrule
        The claim is false because the Amber Alert system was not shut down during the government shutdown. The system is run by individual states and coordinated through a private nonprofit organization, not the federal government. The shutdown only affected the federal webpage promoting the program, not the actual operation of the Amber Alert program. Alerts were still sent out in different states during the shutdown. The federal webpage was purely informational and does not impact the functioning of the program. Thus, the claim that \textcolor{cyan}{the Obama administration shut down the Amber Alert program because of the government shutdown} is not accurate.\\
        \midrule
        \multicolumn{1}{c}{\textbf{Llama-70B ( Zero-shot) Explanation Task}} \\
        \midrule
        The claim that \textcolor{cyan}{the Obama administration shut down the Amber Alert program is false} because the Amber Alert system is not run by the federal government, but rather by individual states and coordinated through a private nonprofit organization. The federal webpage promoting the program was temporarily closed due to the government shutdown, but this did not affect the operation of the program itself. The Amber Alert system continued to operate during the shutdown, with alerts being sent out in Texas and Florida. The federal webpage was restored to eliminate confusion among the public, but it was purely informational and did not impact the functioning of the program. The Amber Alert system is activated by local or state law enforcement officials when they have specific details about a missing child, such as their name, photo, and information about the suspect or vehicle involved. The system relies on media and cell phone notifications to spread the word and has been successful in locating missing children in the past. The federal webpage has never been involved in the actual process of finding missing children. \textcolor{cyan}{Overall, the claim that the Obama administration shut down the Amber Alert program is false} and based on a misunderstanding of how the system operates. \\
        \midrule
        \multicolumn{1}{c}{\textbf{GPT4 (few-shot) Explanation Task}} \\
        \midrule
        The US federal shutdown did not shut down the amber alert system, despite incorrect reports to the contrary. The official US government amber alert website, which promotes the program and educates the public about amber alerts, was taken offline, but this did not affect the functioning of the amber alert system itself. Amber alerts are actually organized, activated, and issued at the state and local level, and they continued to be issued during the shutdown. Amber Alerts are emergency messages that are broadcast when a law enforcement agency determines that a child has been abducted and is in imminent danger. They are broadcast via radio, television, road signs, cell phones, and other data-enabled devices. While it’s important to note that the federal website is not the amber alert system. The federal website promotes the program and issues guidelines, but it has no active role in issuing amber alerts. While it’s also correct to observe that funding for the amber alert program comes from a mixture of federal, state, and local resources. The office of justice programs, which oversees the amber alert program, is not paid directly by the federal government. Instead, its funding flows from the Department of Justice to state and local partners. Therefore, even during a federal shutdown, essential public safety programs such as the amber alert program would continue to operate.\\
        \midrule
        \multicolumn{1}{c}{\textbf{Llama-70B (few-shot) Explanation Task}} \\
        \midrule
        The Amber Alert program was not shut down during the government shutdown.\\
        \midrule
        \multicolumn{1}{c}{\textbf{Vicuna-13B (PEFT) Explanation Task}} \\
        \midrule
        \textcolor{cyan}{The Obama administration shut down the Amber Alert program because of the government shutdown.}
        \\
        \midrule
        \multicolumn{1}{c}{\textbf{GPT4 (zero-shot) Joint Task}} \\
        \midrule
        The context contradicts the claim that \textcolor{cyan}{the Obama administration shut down the Amber Alert program because of the government shutdown}. According to the context, the Amber Alert system is operated by individual states and coordinated through a private nonprofit organization. Although the federal webpage promoting the program was temporarily closed due to the government shutdown, the Amber Alert program itself continued its operation. Thus, the claim is not supported by the context and is labeled as false.\\
\end{tabular}
\caption{Part 1: Example explanations.}
\end{subtable}
\end{table*}
\begin{table*}[t!p]
\ContinuedFloat
\small
\begin{subtable}{\textwidth}
\begin{tabular}{p{15.5cm}}
        \midrule
        \multicolumn{1}{c}{\textbf{Vicuna-13B (zero-shot) Joint Task}} \\
        \midrule
        \textcolor{cyan}{The claim that the Obama administration shut down the Amber Alert program due to the government shutdown is false. The Amber Alert system is not run by the federal government, but rather by individual states and coordinated through the National Center for Missing and Exploited Children, a private nonprofit organization. The federal webpage promoting the program was temporarily closed due to the government shutdown, but this did not affect the operation of the program itself. The Amber Alert system continued to operate during the shutdown, with alerts being sent out in Texas and Florida. The federal webpage was restored to eliminate confusion among the public, but it was purely informational and did not impact the functioning of the program. The Amber Alert system is activated by local or state law enforcement officials when they have specific details about a missing child, such as their name, photo, and information about the suspect or vehicle involved. The system relies on media and cell phone notifications to spread the word and has been successful in locating missing children in the past. The federal webpage has never been involved in the actual process of finding missing children. Overall, the claim that the Obama administration shut down the Amber Alert program is false and based on a misunderstanding of how the system operates.}
        \\
        \midrule
        \multicolumn{1}{c}{\textbf{GPT-3.5-D (few-shot) Joint Task}} \\
        \midrule
        Federal government shutdown did not affect the functioning of the Amber Alert system, which is coordinated through a private nonprofit organization
        \\
        \midrule
        \multicolumn{1}{c}{\textbf{Vicuna-13B (few-shot) Joint Task}} \\
        \midrule
        The Amber Alert system is run by individual states and coordinated through the National Center for Missing and Exploited Children, a private nonprofit organization. The federal government shutdown did not affect the operation of the program, which continued to function during the shutdown. The federal government's webpage promoting the program was temporarily closed, but this did not impact the program's operation. The claim that the Obama administration shut down the Amber Alert program is false.\\
        \midrule
        \multicolumn{1}{c}{\textbf{Vicuna-13B (PEFT) Joint Task}} \\
        \midrule
        What's true: The federal government's website promoting the Amber Alert system was down during the government shutdown. What's false: The Amber Alert system is not run by the federal government, but by individual states and coordinated through a private nonprofit. The government shutdown had no effect on the Amber Alert system. \textcolor{orange}{The ""Share Your Renewal Act"" is unrelated to the Amber Alert system.}
        \\
        \midrule
        \multicolumn{1}{c}{\textbf{Gold Explanation}} \\
        \midrule
        People on Twitter, bloggers and others said the Obama administration shut down the Amber Alert system. Actually, each state has its own alert system, and national coordination takes place through the National Center for Missing and Exploited Children, a private nonprofit, untouched by the federal stalemate. The system has continued to operate through the federal shutdown, sending out alerts in Texas and Florida. The one sliver of truth that triggered this claim is that a federal webpage promoting the program was closed, but that did not affect the program itself. It is a mistake to confuse that web page with the Amber Alert system.
        \\
        \toprule
        \multicolumn{1}{c}{\textbf{Example 2}} \\
        \midrule
        \textbf{Claim:} United States health care facilities face an intravenous bag shortage because of the devastation wreaked by Hurricane Maria in September 2017.
        
        \textbf{Summarized Context:}
        The article discusses the shortage of intravenous (IV) bags in both Puerto Rico and the mainland United States as a result of the damage caused by Hurricane Maria in September 2017. The tweet from Ben Boyer criticizing the Trump administration's response to the disaster brought attention to this issue. The shortage of IV bags is due to the fact that many medical manufacturing plants, including Baxter International, the largest IV bag supplier in the US, are located in Puerto Rico and were affected by the hurricane. The Food and Drug Administration (FDA) has been working closely with industry and officials to address the shortage of IV saline and other products. The FDA expects the shortage of IV fluids to improve in early 2018 based on information from the manufacturers. Baxter International confirmed that its facilities in Puerto Rico are connected to the local electrical grid and expects to return to normal supply levels for products made in Puerto Rico in the coming weeks. However, Baxter clarified that its Puerto Rican operations did not include the manufacture of saline IV solutions, and production of those products continues to run normally. The FDA Commissioner, Scott Gottlieb, stated that addressing the IV saline shortage is a key area of focus and expects the situation to improve early this year. The article also mentions that the number of deaths caused by Hurricane Maria remains unclear, with conflicting reports and a recount of the death toll ordered by the Governor of Puerto Rico. The shortage of IV bags has affected healthcare facilities in the United States, leading to challenges in administering necessary treatments. Boyer's tweet brought attention to this issue, and the story gained traction on social media, with many nurses and medical professionals sharing their experiences with the shortage. In summary, the article highlights the shortage of IV bags in the United States due to the damage caused by Hurricane Maria in Puerto Rico. The FDA and manufacturers are working to address the shortage, and improvements are expected in early 2018. The shortage has impacted healthcare facilities and patients, leading to challenges in providing necessary treatments.
        \\
        \bottomrule         
    \end{tabular}
  \caption{Part 2: Example explanations.}
  \end{subtable}
\end{table*}
\begin{table*}[t!p]
\ContinuedFloat
\small
\begin{subtable}{\textwidth}
\begin{tabular}{p{15.5cm}}
        \midrule
        \multicolumn{1}{c}{\textbf{GPT4 (zero-shot) Explanation Task}} \\
        \midrule
        The context verifies the claim as true due to several reasons. It explains how Hurricane Maria in 2017 caused damage to Puerto Rico, where many medical manufacturing plants, including those from Baxter International (the largest IV bag supplier in the U.S), are situated. This damage led to a shortage of IV bags both in Puerto Rico and mainland United States. The FDA and manufacturers, including Baxter, are working to address this shortage, suggesting that it has had significant effects on healthcare facilities, which is also supported by various personal accounts shared on social media. \textcolor{violet}{However, the impact extends beyond the shortage of IV bags, as the disaster has caused challenges in administering necessary treatments.} Consequently, the claim that U.S healthcare facilities are facing an IV bag shortage due to the devastation caused by Hurricane Maria is true based on the provided context.
        \\
        \midrule
        \multicolumn{1}{c}{\textbf{Llama-70B (zero-shot) Explanation Task}} \\
        \midrule
        The claim that \textcolor{cyan}{United States health care facilities face an intravenous bag shortage because of the devastation wreaked by Hurricane Maria in September 2017} is true. The article states that the shortage of intravenous (IV) bags in both Puerto Rico and the mainland United States is a result of the damage caused by Hurricane Maria in September 2017. The shortage of IV bags is due to the fact that many medical manufacturing plants, including Baxter International, the largest IV bag supplier in the US, are located in Puerto Rico and were affected by the hurricane. The Food and Drug Administration (FDA) has been working closely with industry and officials to address the shortage of IV saline and other products. The FDA expects the shortage of IV fluids to improve in early 2018 based on information from the manufacturers. Baxter International confirmed that its facilities in Puerto Rico are connected to the local electrical grid and expects to return to normal supply levels for products made in Puerto Rico in the coming weeks. However, Baxter clarified that its Puerto Rican operations did not include the manufacture of saline IV solutions, and production of those products continues to run normally. The FDA Commissioner, Scott Gottlieb, stated that addressing the IV saline shortage is a key area of focus and expects the situation to improve early this year. The shortage of IV bags has affected healthcare facilities in the United States, leading to challenges in administering necessary treatments. In summary, the shortage of IV bags in the (\textcolor{red}{Incomplete explanation})
        \\
        \midrule
        \multicolumn{1}{c}{\textbf{GPT4 (few-shot) Explanation Task}} \\
        \midrule
        A months-long IV bag shortage might become a life-or-death situation for thousands of Americans and is almost certainly impeding care at Seattle Children’s Hospital, among other pediatric care facilities around the country, according to reports from health-care professionals. One of the reports was a post from former television executive Ben Boyer that went viral last week. His tweet shows an IV bag on a pole and \textcolor{orange}{was captioned: “My wife’s nurse had to stand for 30 mins \& administer a drug slowly through a syringe because there are almost no IV bags in the continental U.S. anymore. Out of PR Belt [Pharms] makes them. And we’re using [Puerto Rico’s] only other natural resource to ship supplies: people.” Boyer’s wife was identified by several news outlets as Xenocrates Boyer, an experienced Hollywood set designer who has worked on productions as varied as Scandal, Dexter, and American Pie. Ben Boyer is a co-executive producer of the FX series You’re the Worst.} Baxter International’s website lists heparin and potassium chloride as two drugs it manufacturers in IV solution form. Earlier this month, Bloomberg reported that millions of gallons of intravenous solutions were available but inaccessible because federal agencies were wrestling with how to get the fluids off the island and onto the mainland. Baxter says on its website that several of its Puerto Rican manufacturing locations were damaged in the hurricane and the company is having to air-freight products from less storm-torn areas to fill the (\textcolor{red}{Incomplete explanation})
        \\
        \midrule
        \multicolumn{1}{c}{\textbf{Llama-70B (few-shot) Explanation Task}} \\
        \midrule
        This story explains that a shortage of IV bags in both Puerto Rico and the mainland United States is due to the damage caused by Hurricane Maria in September 2017. The story does a good job of explaining that the shortage is due to the fact that many medical manufacturing plants, including Baxter International, the largest IV bag supplier in the United States, are located in Puerto Rico and were affected by the hurricane. The story also includes a quote from the FDA Commissioner, Scott Gottlieb, who stated that addressing the IV saline shortage is a key area of focus and expects the situation to improve early this year.\\
        \midrule
        \multicolumn{1}{c}{\textbf{Vicuna-13B (PEFT) Explanation Task}} \\
        \midrule
        \textcolor{cyan}{Because of Hurricane Maria, the United States is facing a shortage of intravenous (IV) bags.}\\
        \midrule
        \multicolumn{1}{c}{\textbf{GPT4 (zero-shot) Joint Task}} \\
        \midrule
        The context supports the claim that United States health care facilities face an intravenous bag shortage as a result of the damage caused by Hurricane Maria in Puerto Rico in September 2017. The context mentions that the leading IV bag supplier in the US, Baxter International, has operations in Puerto Rico and these were affected by the hurricane. The disruption contributed to a shortage of IV bags, which has impacted health care provision in the United States. The Food and Drug Administration (FDA) is working with manufacturers to try and alleviate the shortage.
        \\
        \midrule
        \multicolumn{1}{c}{\textbf{Vicuna-13B (zero-shot) Joint Task}} \\
        \midrule
        The claim is true as the article discusses the shortage of intravenous (IV) bags in both Puerto Rico and the mainland United States as a result of the damage caused by Hurricane Maria in September 2017. The shortage of IV bags is due to the fact that many medical manufacturing plants, including Baxter International, the largest IV bag supplier in the US, are located in Puerto Rico and were affected by the hurricane.
        \\
        \midrule
        \multicolumn{1}{c}{\textbf{GPT-3.5-D (few-shot) Joint Task}} \\
        \midrule
        The article states that the shortage of IV bags is due to the damage caused by Hurricane Maria in Puerto Rico, with healthcare facilities facing challenges in providing treatments.\\
        \bottomrule         
    \end{tabular}
  \caption{Part 3: Example explanations.}
  \end{subtable}
\end{table*}
\begin{table*}[t!p]
\ContinuedFloat
\small
\begin{subtable}{\textwidth}
\begin{tabular}{p{15.5cm}}
        \midrule
        \multicolumn{1}{c}{\textbf{Vicuna-13B (few-shot) Joint Task}} \\
        \midrule
        The article discusses the shortage of IV bags in the United States due to the damage caused by Hurricane Maria in Puerto Rico. The FDA and manufacturers are working to address the shortage, and improvements are expected in early 2018. The shortage has impacted healthcare facilities and patients, leading to challenges in providing necessary treatments.
        \\      
        \midrule
        \multicolumn{1}{c}{\textbf{Vicuna-13B (PEFT) Joint Task}} \\
        \midrule
        \textcolor{orange}{We contacted Baxter International to ask about the status of its IV bag production in Puerto Rico, but have not yet received a response.} However, the FDA has been working with Baxter and other manufacturers to address the shortage, and expects the situation to improve in early 2018.
        \\
        \midrule
        \multicolumn{1}{c}{\textbf{Gold Explanation}} \\
        \midrule
        \textcolor{orange}{Most of the people who have contacted him since his remark began spreading online, he said, have offered similar stories.}\\
        \toprule
        \multicolumn{1}{c}{\textbf{Example 3}} \\
        \midrule
        \textbf{Claim:} The new supplement InteliGEN can boost brain function.
        
        \textbf{Summarized Context:} The article provided does not contain any information about the claim regarding the supplement InteliGEN and its ability to boost brain function. Therefore, it is not possible to summarize the article in relation to the claim.
        \\
        \midrule
        \multicolumn{1}{c}{\textbf{Vicuna-13B (PEFT) Explanation Task}} \\
        \midrule
        \textcolor{orange}{A 7.5 magnitude earthquake struck off the coast of Indonesia, killing at least 91 people, injuring hundreds and damaging hundreds of buildings, including the country’s biggest mosque.}\\
        \midrule
        \multicolumn{1}{c}{\textbf{Vicuna-13B (PEFT) Joint Task}} \\
        \midrule
        \textcolor{orange}{Aid groups in South Africa’s Gauteng province are struggling to contain an outbreak of measles among refugees, highlighting the vulnerability of migrants who often live in crowded conditions and have little access to healthcare.}\\
        \midrule
        \multicolumn{1}{c}{\textbf{Gold Explanation}} \\
        \midrule
        \textcolor{orange}{Tens of thousands of holiday makers fled seaside towns on Australia’s east coast on Thursday as bushfires approached, and military ships and helicopters began rescuing thousands more trapped by the blazes.}\\
        
        \bottomrule         
    \end{tabular}
  \caption{Part 4: Example explanations.}
  \end{subtable}

  \caption{The generated explanations of different LLMs. \textcolor{cyan}{Claim repetition, internal repetition, or copy context as explanation}. \textcolor{orange}{Extra Information}. \textcolor{violet}{External Inconsistency}. During the human evaluation of gold explanations, we assess the criteria of Extra Information, External Consistency, and Missing Information with respect to the original context, not the summarized context. For the third example, we only include models that struggle with unproven claims, generating irrelevant text as explanation, while excluding models that produce acceptable explanations.}    
  \label{generated-examples}
\end{table*}

\begin{table*}[ht!]
\centering
\resizebox{\textwidth}{!}
    {
    \begin{tabular}{clccccccc}
        \toprule
        \multirow{3}{*}{\textbf{\rotatebox[origin=c]{90}{\small Setting}}} &
        \multirow{3}{*}{\textbf{Model}} & \multicolumn{7}{c}{\textbf{Veracity Task  / Joint Task}} \\
        \cmidrule(lr){3-9}
        & & \multicolumn{3}{c}{\textbf{Macro}} & \multicolumn{3}{c}{\textbf{Weighted}} & \textbf{Acc.}\\
        \cmidrule(lr){3-5}
        \cmidrule(lr){6-8}
        & & \textbf{Pr.} & \textbf{Rc.} & \textbf{F1.}& \textbf{Pr.} & \textbf{Rc.} & \textbf{F1.} &  \\ 
          \midrule
          
          & Majority & 12.2 & 25.0 & 16.4 & 23.6 & 48.6 & 31.8 & 48.6\\
         \midrule
         \multirow{7}{*}{\rotatebox[origin=c]{90}{\textbf{\small Zero-shot}}}
          & GPT-3.5-D & 54.1 / 52.2 & 55.6 / 54.0 & 51.7 / 50.0 & 75.1 / 73.6 & 63.8 / 61.4 & 67.8 / 65.9 & 63.8 / 61.4 \\
          & GPT-3.5-T & 56.0 / 53.3 & 52.6 / 55.6 & 51.4 / \textbf{53.9}
          & 76.5 / 76.0 & 66.80 / 67.8 & 69.3 / \textbf{70.7} & 66.80 / 67.8\\
          
          & GPT-4 & 54.3 / 54.3 & 55.4 / 55.8 & \textbf{53.2} / 53.4 & 73.4 / 73.3 & 67.5 / 67.3 & \textbf{69.8} / 69.6 & 67.5/ 67.3 \\          
          \cmidrule(lr){2-9}        
          & Falcon-180B & 59.6 / 63.4 & 39.9 / 47.1 & \textbf{36.6} / 44.2 & 70.9 / 73.5 & 66.7 / 73.9 & \textbf{59.0} / \textbf{66.6} & 66.7 / 73.9 \\
          
          & Llama-70B & 43.9 / 34.0 & 37.2 / 34.0 & 33.8 / 31.2 & 58.0 / 50.0 & 53.2 / 49.1 & 49.4 / 46.2 & 53.2 / 49.1 \\
                
          & Vicuna-13B & 57.0 / 57.8 & 34.6 / 49.6 & 23.2 / \textbf{47.4} & 70.7 / 75.8 & 29.5 / 58.0 & 24.5 / 61.4 & 29.5 / 58.0\\
          
          & Mistral-7B & 51.4 / 46.7 & 28.5 / 46.0 & 20.5 / 41.5 & 72.9 / 68.1 & 25.7 / 49.8 & 25.0 / 55.5 & 25.7 / 49.8\\          
         \midrule
         
         \multirow{7}{*}{\rotatebox[origin=c]{90}{\textbf{\small Few-shot}}}
          & GPT-3.5-D [4/1] & 50.6 / 57.0 & 51.2 / 56.7 & 49.9 / \textbf{56.6} & 68.0 / 73.7 & 68.3 / 72.3 & 67.7 / \textbf{72.9} & 68.3 / 72.3\\

          & GPT-3.5-T [2/7] & 54.5 / 55.7 & 53.7 / 55.3 & 52.9 / 54.5 & 74.6 / 67.3 & 67.8 / 69.8 & \textbf{70.1} / 67.5 & 67.8 / 69.8\\

          & GPT-4 [2/9] & 54.9 / 55.5 & 56.1 / 56.9 & \textbf{53.0} / 54.9 & 75.0 / 74.1 & 66.2 / 70.0 & 69.7 / 71.5 & 66.2 / 70.0 \\

          \cmidrule(lr){2-9}     
          & Falcon-180B [2/1] & 57.7 / 54.8 & 58.9 / 52.3 & \textbf{57.9} / 51.2 & 75.6 / 73.3 & 74.0 / 68.8 & \textbf{74.8} / 70.0 & 74.0 / 68.8 \\

          & Llama-70B [4/4] & 52.5 / 50.8 & 52.0 / 53.2 & 49.3 / 49.0 & 71.1 / 76.6 & 68.8 / 70.3 & 68.6 / 72.6 & 68.8 / 70.3 \\

          & Vicuna-13B [6/7] & 52.2 / 55.8 & 53.8 / 56.0 & 52.4 / \textbf{54.8} & 72.0 / 76.6 & 68.1 / 74.1 & 69.7 / \textbf{75.0} & 68.1 / 74.1\\

          & Mistral-7B [9/6] & 59.5 / 51.9 & 48.8 / 64.1 & 44.9 / 51.6 & 75.2 / 89.5 & 73.6 / 76.5 & 67.9 / 81.8 & 73.6 / 76.5\\
         \midrule
         \multirow{2}{*}{\rotatebox[origin=c]{90}{\textbf{\small PEFT}}}
      
          & Vicuna-13B & 69.7 / 71.6 & 67.8 / 68.9 & 68.5 / 70.0 & 80.9 / 81.3 & 80.4 / 81.2 & 80.5 / 81.2 & 80.4 / 81.2 \\
          
          & Mistral-7B & 75.5 / 74.2 & 70.3 / 68.4 & \textbf{72.0} / \textbf{70.1} & 82.9 / 82.6 & 82.3 / 81.8 & \textbf{82.5} / \textbf{82.0} & 82.3 / 81.8 \\
        \bottomrule
    \end{tabular}
    }
  \caption{Veracity prediction results on the test set. The models' performance is evaluated using precision (Pr.), recall (Rc.), F1, and accuracy (Acc.) metrics.}
  \label{veracity-detailed-results} 
\end{table*}

\begin{table*}[ht!]
    \resizebox{\textwidth}{!}
    {
    \begin{tabular}{cllcccccc}
        \toprule

        \multirow{2}{*}{\textbf{\rotatebox[origin=c]{-90}{Setting}}} &
        \multirow{2}{*}{\textbf{Model}} &
        \multirow{2}{*}{\textbf{Evaluation Method}} &
        \multicolumn{3}{c}{\textbf{Explanation Task}} &
        \multicolumn{3}{c}{\textbf{Joint Task}}\\
        \cmidrule(lr){4-6}
        \cmidrule(lr){7-9}
         & & &
        \textbf{SGC} & \textbf{WGC} & \textbf{LC} &
        \textbf{SGC} & \textbf{WGC} & \textbf{LC} \\
         \midrule
         & \multirow{4}{*}{Gold Explanations} & DA+ELMO:SNLI & - & -& - & 25.0 & 82.79 & 63.31\\
         & & RoBERTa:SNLI & - & -& - & 22.32 & 78.17 & 57.87 \\
         & & RoBERTa:MNLI & - & -& - & 22.24 & 90.83 & 70.29\\
         & & Roberta-L:(S+M+A)NLI-FEVER & - & -& - & 22.0 & 93.02 & 75.24\\
         \midrule   
         \multirow{28}{*}{\textbf{\rotatebox[origin=c]{-90}{Zero-shot}}}
          & \multirow{4}{*}{GPT-3.5-D} & DA+ELMO:SNLI & 9.5 & 75.32 & 46.43 & 19.24 & 81.41 & 63.39\\
         & & RoBERTa:SNLI & 3.57 & 72.89 & 40.18 & 13.31 & 76.87 & 59.25\\
         & & RoBERTa:MNLI & 2.92 & 87.99 & 80.03 & 11.61 & 87.74 & 89.2\\
         & & Roberta-L:(S+M+A)NLI-FEVER & 3.73 & 90.18 & 87.66 & 12.66 & 90.67 & 90.91 \\
         \cmidrule(lr){2-9}
         
          & \multirow{4}{*}{GPT-3.5-T} & DA+ELMO:SNLI & 2.52 & 66.96 & 10.63 & 2.6 & 66.88 & 27.84\\
         & & RoBERTa:SNLI & 0.08 & 57.95 & 12.99 & 1.06 & 59.01 & 29.22 \\
         & & RoBERTa:MNLI & 0.24 & 83.6 & 55.36 & 1.22 & 86.04 & 78.25 \\
         & & Roberta-L:(S+M+A)NLI-FEVER & 0.24 & 84.21 & 42.11 & 0.97 & 88.88 & 81.9 \\
         \cmidrule(lr){2-9}
         
          & \multirow{4}{*}{GPT-4} & DA+ELMO:SNLI & 4.95 & 72.16 & 29.95 & 8.12 & 75.24 & 40.75\\
         & & RoBERTa:SNLI & 1.70 & 68.26 & 28.98 & 5.28 & 70.45 & 42.53\\
         & & RoBERTa:MNLI & 2.03 & 87.42 & 71.75 & 5.11 & 88.8 & 83.6\\
         & & Roberta-L:(S+M+A)NLI-FEVER & 1.41 & 92.74 & 81.03 & 5.19 & 90.58 & 87.5 \\
         \cmidrule(lr){2-9}

        & \multirow{4}{*}{Falcon-180B} & DA+ELMO:SNLI & 7.79 & 64.61 & 23.86 & 5.6 & 64.61 &  52.35 \\
         & & RoBERTa:SNLI & 4.79 & 57.71 & 21.67 & 3.17 & 65.34 & 51.95\\
         & & RoBERTa:MNLI & 4.46 & 77.6 & 50.49 & 2.6 & 89.37 & 75.41 \\
         & & Roberta-L:(S+M+A)NLI-FEVER & 4.38 & 81.09 & 57.95 & 2.52 & 91.23 & 78.49\\
         \cmidrule(lr){2-9}
         
          & \multirow{4}{*}{Llama-70B} & DA+ELMO:SNLI & 9.12 & 69.59 & 29.11 & 7.71 & 73.33 & 29.21 \\
         & & RoBERTa:SNLI & 4.43 & 62.29 & 26.59 & 4.83 & 65.28 & 25.99\\
         & & RoBERTa:MNLI & 4.17 & 81.06 & 56.99 & 3.98 & 79.93 & 70.87\\
         & & Roberta-L:(S+M+A)NLI-FEVER & 4.0 & 83.75 & 61.77 & 4.06 & 82.73 & 76.29 \\
         \cmidrule(lr){2-9}
         
          & \multirow{4}{*}{Vicuna-13B} & DA+ELMO:SNLI & 1.54 & 59.01 & 6.57 & 2.52 & 65.75 & 24.11 \\
         & & RoBERTa:SNLI & 0.0 & 50.65 & 4.3 & 1.06 & 60.8 & 23.54\\
         & & RoBERTa:MNLI & 0.0 & 74.59 & 34.25 & 0.81 & 81.9 & 61.61\\
         & & Roberta-L:(S+M+A)NLI-FEVER & 0.0 & 78.49 & 44.89 & 0.81 & 86.61 & 66.8  \\
         \cmidrule(lr){2-9}
         
          & \multirow{4}{*}{Mistral-7B} & DA+ELMO:SNLI & 1.79 & 57.55 & 6.49 & 2.92 & 61.69 & 13.96\\
         & & RoBERTa:SNLI & 0.24 & 50.41 & 4.87 & 01.06 & 53.17 & 15.18\\
         & & RoBERTa:MNLI & 0.24 & 70.94 & 27.52 & 0.89 & 78.33 & 49.84\\
         & & Roberta-L:(S+M+A)NLI-FEVER & 0.24 & 75.0 & 33.93 & 0.89 & 81.74 & 55.44 \\
         \midrule
         
         \multirow{16}{*}{\textbf{\rotatebox[origin=c]{-90}{Few-shot}}}
          & \multirow{4}{*}{GPT-3.5-D [1/1]} & DA+ELMO:SNLI & 7.63 & 73.54 & 38.23 & 36.93 & 90.34 & 94.89 \\
         & & RoBERTa:SNLI & 2.27 & 67.05 & 33.2 & 29.14 & 91.15 & 92.53 \\
         & & RoBERTa:MNLI & 1.95 & 86.44 & 81.33 & 27.92 & 92.86 & 97.4 \\
         & & Roberta-L:(S+M+A)NLI-FEVER & 2.19 & 89.29 & 84.42 & 28.41 & 93.75 & 98.62 \\
         \cmidrule(lr){2-9}
         
          & \multirow{4}{*}{GPT-3.5-T [5/7]} & DA+ELMO:SNLI & 7.63 & 74.11 & 29.87 & 22.89 & 84.5 & 67.13 \\
         & & RoBERTa:SNLI & 2.68 & 70.62 & 29.71 & 16.72 & 81.98 & 64.12 \\
         & & RoBERTa:MNLI & 2.84 & 88.88 & 73.94 & 15.91 & 90.34 & 91.64 \\
         & & Roberta-L:(S+M+A)NLI-FEVER & 2.52 & 90.02 & 79.87 & 15.99 & 91.4 & 93.99 \\
         \cmidrule(lr){2-9}
         
          & \multirow{4}{*}{GPT-4 [11/9]} & DA+ELMO:SNLI & 19.89 & 79.22 & 60.55 & 18.02 & 81.09 & 59.74 \\
         & & RoBERTa:SNLI & 14.77 & 76.54 & 57.31 & 13.23 & 77.76 & 55.93 \\
         & & RoBERTa:MNLI & 14.2 & 89.04 & 78.25 & 13.31 & 88.56 & 84.66 \\
         & & Roberta-L:(S+M+A)NLI-FEVER & 15.26 & 90.58 & 82.63 & 13.64 & 91.31 & 89.04 \\
         \cmidrule(lr){2-9}
         
         & \multirow{4}{*}{Falcon-180B [1/1]} & DA+ELMO:SNLI & 00.24 & 53.98 & 13.47 & 00.57 & 56.01 & 18.18 \\
         & & RoBERTa:SNLI & 00.00 & 49.19 & 09.09 & 00.08 & 45.94 & 12.58 \\
         & & RoBERTa:MNLI & 00.00 & 81.33 & 32.79 & 00.08 & 80.84 & 37.74 \\
         & & Roberta-L:(S+M+A)NLI-FEVER & 00.00 & 80.84 & 39.45 & 00.08 & 81.17 & 44.56 \\
        \bottomrule        
    \end{tabular}
    }
  \caption{The NLI-based coherence metrics on the test set for explanation generation and the joint task using different NLI models (part one).}
  \label{coherence-full-results}
\end{table*}

\begin{table*}[ht!]
    \resizebox{\textwidth}{!}
    {
    \begin{tabular}{cllcccccc}
        \toprule

        \multirow{2}{*}{\textbf{\rotatebox[origin=c]{-90}{Setting}}} &
        \multirow{2}{*}{\textbf{Model}} &
        \multirow{2}{*}{\textbf{Evaluation Method}} &
        \multicolumn{3}{c}{\textbf{Explanation Task}} &
        \multicolumn{3}{c}{\textbf{Joint Task}}\\
        \cmidrule(lr){4-6}
        \cmidrule(lr){7-9}
         & & &
        \textbf{SGC} & \textbf{WGC} & \textbf{LC} &
        \textbf{SGC} & \textbf{WGC} & \textbf{LC} \\
         \midrule
         \multirow{12}{*}{\textbf{\rotatebox[origin=c]{-90}{Few-shot}}}      
          & \multirow{4}{*}{Llama-70B [4/4]} & DA+ELMO:SNLI & 33.22 & 82.63 & 68.38 & 21.51 & 72.16 & 42.86 \\
         & & RoBERTa:SNLI & 31.45 & 82.04 & 65.35 & 20.78 & 66.72 & 37.5 \\
         & & RoBERTa:MNLI & 30.44 & 92.5 & 78.84 & 20.62 & 87.34 & 57.63 \\
         & & Roberta-L:(S+M+A)NLI-FEVER & 30.35 & 94.27 & 81.11 & 20.54 & 88.64 & 64.37 \\
         \cmidrule(lr){2-9}
         
          & \multirow{4}{*}{Vicuna-13B [5/7]} & DA+ELMO:SNLI & 2.27 & 59.09 & 9.74 & 11.77 & 71.27 & 34.33 \\
         & & RoBERTa:SNLI & 1.3 & 49.27 & 6.74 & 7.87 & 66.31 & 30.44 \\
         & & RoBERTa:MNLI & 0.97 & 75.57 & 36.12 & 7.87 & 81.57 & 61.28 \\
         & & Roberta-L:(S+M+A)NLI-FEVER & 0.97 & 78.73 & 43.51 & 7.39 & 85.96 & 70.86 \\
         \cmidrule(lr){2-9}
         
          & \multirow{4}{*}{Mistral-7B [3/6]} & DA+ELMO:SNLI & 13.64 & 74.35 & 47.16 & 7.87 & 63.96 & 18.59 \\
         & & RoBERTa:SNLI & 9.01 & 68.43 & 43.75 & 7.14 & 56.49 & 16.15 \\
         & & RoBERTa:MNLI & 8.36 & 84.58 & 65.99 & 7.14 & 79.3 & 37.66 \\
         & & Roberta-L:(S+M+A)NLI-FEVER & 8.85 & 87.18 & 71.27 & 7.06 & 83.77 & 45.37 \\
         \midrule
         
         \multirow{8}{*}{\textbf{\rotatebox[origin=c]{-90}{PEFT}}}
          & \multirow{4}{*}{Vicuna-13B} & DA+ELMO:SNLI & 32.63 & 85.63 & 68.34 &  27.35 & 84.33 & 64.04 \\
         & & RoBERTa:SNLI & 31.09 & 82.06 & 63.39 & 25.49 & 79.79 & 60.47 \\
         & & RoBERTa:MNLI & 30.60 & 92.45 & 72.65 & 23.86 & 91.48 & 70.70 \\
         & & Roberta-L:(S+M+A)NLI-FEVER & 30.52 & 93.99 & 75.57 & 25.00 & 92.69 & 73.54 \\
         \cmidrule(lr){2-9}
         
          & \multirow{4}{*}{Mistral-7B} & DA+ELMO:SNLI & 26.79 & 85.06 & 67.29 & 30.28 & 85.55 & 67.94 \\
         & & RoBERTa:SNLI & 22.89 & 79.06 & 61.53 & 27.84 & 81.66 & 63.64\\
         & & RoBERTa:MNLI & 23.13 & 91.15 & 72.40 & 27.19 & 91.88 & 73.78 \\
         & & Roberta-L:(S+M+A)NLI-FEVER & 23.13 & 93.18 & 75.89 & 26.7 & 92.21 & 76.7 \\
        \bottomrule        
    \end{tabular}
    }
  \caption{The NLI-based coherence metrics on the test set for explanation generation and the joint task using different NLI models (part two).}    
  \label{coherence-full-results2}
\end{table*}

\begin{table*}[ht!]
\centering
    \setlength{\tabcolsep}{15pt}
    \begin{tabular}{ccccccc}
        \toprule
        \textbf{C. Rep.} &
        \textbf{I. Rep.} &
        \textbf{S. class F1} &
        \textbf{I. Cons.} &
        \textbf{E. Cons.} &
        \textbf{Extra} &
        \textbf{Missing}\\ 
        \midrule

        0.82 & 0.96 &
        0.84 &
        0.88 &
        0.82 &
        0.94 &
        0.71 \\
        \bottomrule 
    \end{tabular}
    \caption{Agreement percentages across various criteria in the human evaluation process.}
    \label{human_eval-agreement_table}

\end{table*}

\begin{table*}
\begin{subtable}[c]{0.5\textwidth}
\centering
\tiny
    \setlength{\tabcolsep}{1pt}
    \begin{tabular}{c|cccccccccc}
        \toprule
        \textbf{} &
        \textbf{C. Rep.} &
        \textbf{I. Rep.} &
        \textbf{S. class F1} &
        \textbf{I. Cons.} &
        \textbf{E. Cons.} &
        \textbf{Extra} &
        \textbf{Missing}&
        \textbf{S3} &
        \textbf{S5} &
        \textbf{S7} \\ 
        \midrule

        SGC & 0.111 & -0.093 & 0.047 & -0.074 & -0.119 & -0.007 & -0.137 & -0.077 & -0.075 & -0.064 \\
WGC & 0.032 & 0.009 & -0.067 & 0.014 & 0.025 & 0.01 & 0.02 & 0.060 & 0.055 & 0.20 \\
LC & 0.069 & 0.038 & 0.017 & -0.004 & -0.075 & 0.001 & -0.13 & 0.003 & -0.006 & -0.027 \\
R1 & 0.17 & 0.03 & -0.088 & 0.086 & 0.166 & 0.039 & 0.189 & 0.078 & 0.097 & -0.036 \\
R2 & 0.177 & -0.014 & -0.139 & 0.082 & 0.116 & 0.028 & 0.132  & 0.098 & 0.120 & -0.007 \\
RL & 0.236 & 0.012 & -0.133 & 0.098 & 0.132 & 0.007 & 0.133 & 0.121 & 0.138 & -0.043 \\
\end{tabular}
\subcaption{Explanation}
\end{subtable}
\begin{subtable}[c]{0.5\textwidth}
\centering
\tiny
    \setlength{\tabcolsep}{1pt}
    \begin{tabular}{c|cccccccccc}
        \toprule
        \textbf{} &
        \textbf{C. Rep.} &
        \textbf{I. Rep.} &
        \textbf{S. class F1} &
        \textbf{I. Cons.} &
        \textbf{E. Cons.} &
        \textbf{Extra} &
        \textbf{Missing}&
        \textbf{S3} &
        \textbf{S5} &
        \textbf{S7} \\ 
        \midrule
        SGC & 0.142 & -0.05 & -0.019 & 0.014 & 0.013 & 0.021 & 0.025 & 0.089 & 0.090 & -0.016 \\
WGC & -0.041 & -0.041 & -0.105 & -0.042 & -0.052 & 0.059 & -0.076 & 0.091&0.084 &0.055 \\
LC & -0.052 & -0.049 & 0.223 & -0.027 & -0.041 & 0.05 & -0.013 &-0.035 &-0.034 & -0.010 \\
R1 & 0.154 & -0.033 & -0.019 & -0.013 & 0.003 & -0.049 & 0.044 & 0.040 & 0.049 & -0.011  \\
R2 & 0.199 & -0.048 & -0.054 & -0.052 & 0.024 & -0.067 & 0.057 & 0.017 & 0.029 & -0.069 \\
RL & 0.168 & -0.079 & -0.02 & -0.019 & 0.041 & -0.053 & 0.012 &0.053 &0.062 &-0.014 \\
\end{tabular}
\subcaption{Joint}
\end{subtable}
\caption{Correlations between automated metrics and results of the human evaluation. The correlations remained consistently weak, confirming the results of previous work \cite{Local-Interpretations}.}
    \label{correlation-result-table}
\end{table*}

\begin{table*}[ht!]
\centering
\resizebox{\textwidth}{!}
    {
    \setlength{\tabcolsep}{7pt}
    \begin{tabular}{cllrrrrrrrrrr}
        \toprule
        &
        \textbf{Setting} &
        \textbf{Model} &
        \textbf{C. Rep.} &
        \textbf{I. Rep.} &
        \textbf{S. class F1} &
        \textbf{I. Cons.} &
        \textbf{E. Cons.} &
        \textbf{Extra} &
        \textbf{Missing} &
        \textbf{S3} &
        \textbf{S5} &
        \textbf{S7}\\ 
        \midrule
         \multirow{5}{*}{\rotatebox[origin=c]{90}{\textbf{Explanation}}} & \multirow{2}{*}{Zero-shot} &

          GPT-4 & 44.23 & 17.31 & 78.79 & 00.19 & 00.21 & 03.85 & 00.12 & \textbf{76.92} & \textbf{73.08} & 36.54
 \\

           & & Llama-70B & 44.23 & 17.31 & 77.13 & 00.31 & 00.37 & 03.85 & 00.42 & 65.38 & 65.38 & 23.08
  \\
          \cmidrule(lr){2-13}  
        
          & \multirow{2}{*}{Few-shot} &

          GPT-4 & 05.77 & 01.92 & 56.75 & 00.04 & 00.40 & 38.46 & 00.56 & 42.31 & 42.31 & \textbf{38.46}
 \\

            & & Llama-70B & 30.77 & 01.92 & 45.68 & 00.04 & 00.31 & 05.77 & 00.96 & 32.69 & 32.69 & 25.0

 \\
         \cmidrule(lr){2-13}

          & PEFT &
          
          Vicuna-13B & 23.08 & 09.62 & 50.61 & 00.04 & 00.25 & 25.00 & 00.63 & 36.54 & 36.54 & 25.0

 \\
         \midrule
         
         \multirow{5}{*}{\rotatebox[origin=c]{90}{\textbf{Joint}}} & \multirow{2}{*}{Zero-shot} &

           GPT-4 & 32.69 & 03.85 & 53.03 & 00.04 & 00.06 & 03.85 & 00.08 & 59.62 & 57.69 & 38.46

 \\

           & & Vicuna-13B & 38.46 & 11.54 & 48.73 & 00.17 & 00.12 & 03.85 & 00.06 & 55.77 & 51.92 & 25.0

\\
          \cmidrule(lr){2-13}  
          
          & \multirow{2}{*}{Few-shot} &
 GPT-3.5-D & 03.85 & 00.00 & 50.52 & 00.00 & 00.06 & 00.00 & 00.46 & 51.92 & 51.92 & \textbf{48.08}

 \\

           & & Vicuna-13B & 19.23 & 03.85 & 58.25 & 00.04 & 00.00 & 00.00 & 00.21 & \textbf{67.31} & \textbf{67.31} & \textbf{48.08}

  \\
         \cmidrule(lr){2-13}

          & PEFT &
          Vicuna-13B & 09.62 & 07.69 & 64.59 & 00.00 & 00.19 & 36.54 & 00.38 & 42.31 & 42.31 & 40.38

          \\
          \cmidrule(lr){2-13}

          & Gold Exp. & & 07.69 & 00.00 & 41.72 & 00.08 & 00.17 & 50.00 & 00.48 & 25.00 & 25.00 & 19.23\\
        \bottomrule
    \end{tabular}
    }
  \caption{Human evaluation results for the 10 selected models: C. Rep. represents the percentage error of Claim Repetition, I. Rep. signifies the percentage error of Internal Repetition, S. Class F1 denotes the Suggested Class F1, I. Cons. stands for the mean absolute error of Internal Consistency, E. Cons. indicates the mean absolute error of External Consistency, Extra represents the percentage error of Extra Information, and Missing denotes the mean absolute error of Missing Information.}
  \label{human-evaluation-detailed-results} 
\end{table*}

\begin{table*}[ht!]
\centering
\resizebox{\textwidth}{!}
    {
    \setlength{\tabcolsep}{7pt}
\begin{tabular}{lp{7em}p{7em}>{\raggedleft\arraybackslash}p{5em}>{\raggedleft\arraybackslash}p{5em}>{\centering\arraybackslash}p{5em}>{\centering\arraybackslash}p{5em}>{\centering\arraybackslash}p{5em}>{\centering\arraybackslash}p{5em}}
\toprule
& & & & & \multicolumn{4}{c}{\textbf{Prediction}} \\
\cmidrule{6-9}
& \textbf{Setting} & \textbf{Model} & & & True & False & Mixture & Unproven\\
\midrule
\multirow{20}{*}{\rotatebox[origin=c]{90}{\textbf{Explanation}}} & \multirow{8}{*}{Zero-shot} & \multirow{4}{*}{GPT-4} & \multirow{4}{*}{\rotatebox[origin=c]{90}{\textbf{Truth}}} & True & 31 & 0 & 0 & 0\\
& & & & False & 0 & 11 & 1 & 2 \\
& & & & Mixture & 1 & 0 & 3 & 0 \\
& & & & Unproven & 0 & 1 & 0 & 2 \\
\cmidrule{3-9}
& & \multirow{4}{*}{Llama-70B} & \multirow{4}{*}{\rotatebox[origin=c]{90}{\textbf{Truth}}} & True & 28 & 0 & 0 & 3 \\
& & & & False & 0 & 10 & 0 & 4\\
& & & & Mixture & 1 & 0 & 3 & 0\\
& & & & Unproven & 0 & 0 & 0 & 3\\
\cmidrule{2-9}
& \multirow{8}{*}{Few-shot} & \multirow{4}{*}{GPT-4} & \multirow{4}{*}{\rotatebox[origin=c]{90}{\textbf{Truth}}} & True & 26 & 0 & 1 & 4 \\
& & & & False & 1 & 8 & 4 & 1\\
& & & & Mixture & 1 & 0 & 1 & 2\\
& & & & Unproven & 0 & 0 & 0 & 3\\
\cmidrule{3-9}
& & \multirow{4}{*}{Llama-70B} & \multirow{4}{*}{\rotatebox[origin=c]{90}{\textbf{Truth}}} & True & 24 & 0 & 1 & 6\\
& & & & False & 2 & 7 & 2 & 3\\
& & & & Mixture & 3 & 0 & 1 & 0\\
& & & & Unproven & 1 & 1 & 0 & 1\\
\cmidrule{2-9}
& \multirow{4}{*}{PEFT} & \multirow{4}{*}{Vicuna-13B} & \multirow{4}{*}{\rotatebox[origin=c]{90}{\textbf{Truth}}} & True & 27 & 0 & 2 & 2\\
& & & & False & 2 & 6 & 2 & 4\\
& & & & Mixture & 1 & 0 & 2 & 1\\
& & & & Unproven & 1 & 1 & 0 & 1\\
\midrule

\multirow{20}{*}{\rotatebox[origin=c]{90}{\textbf{Joint}}} & \multirow{8}{*}{Zero-shot} & \multirow{4}{*}{GPT-4} & \multirow{4}{*}{\rotatebox[origin=c]{90}{\textbf{Truth}}} & True & 22 & 0 & 6 & 3 \\
& & & & False & 0 & 10 & 3 & 1 \\
& & & & Mixture & 1 & 0 & 1 & 2 \\
& & & & Unproven & 0 & 1 & 0 & 2 \\
\cmidrule{3-9}
& & \multirow{4}{*}{Vicuna-13B} & \multirow{4}{*}{\rotatebox[origin=c]{90}{\textbf{Truth}}} & True & 22 & 2 & 5 & 2 \\
& & & & False & 1 & 6 & 7 & 0 \\
& & & & Mixture & 1 & 0 & 3 & 0 \\
& & & & Unproven & 0 & 2 & 0 & 1 \\
\cmidrule{2-9}
& \multirow{8}{*}{Few-shot} & \multirow{4}{*}{GPT-3.5-D} & \multirow{4}{*}{\rotatebox[origin=c]{90}{\textbf{Truth}}} & True & 27 & 0 & 2 & 2 \\
& & & & False & 2 & 9 & 2 & 1 \\
& & & & Mixture & 2 & 0 & 0 & 2 \\
& & & & Unproven & 0 & 1 & 0 & 2 \\
\cmidrule{3-9}
& & \multirow{4}{*}{Vicuna-13B} & \multirow{4}{*}{\rotatebox[origin=c]{90}{\textbf{Truth}}} & True & 27 & 1 & 2 & 1 \\
& & & & False & 0 & 11 & 2 & 1 \\
& & & & Mixture & 1 & 1 & 0 & 2 \\
& & & & Unproven & 0 & 0 & 0 & 3 \\
\cmidrule{2-9}
& \multirow{4}{*}{PEFT} & \multirow{4}{*}{Vicuna-13B} & \multirow{4}{*}{\rotatebox[origin=c]{90}{\textbf{Truth}}} & True & 25 & 0 & 4 & 2 \\
& & & & False & 1 & 9 & 4 & 0 \\
& & & & Mixture & 1 & 0 & 3 & 0 \\
& & & & Unproven & 0 & 1 & 0 & 2 \\
\bottomrule
\end{tabular}}
\caption{Confusion matrices for the suggested class criterion of the human evaluation.}
\label{confusion_matrix_table} 
\end{table*}

\begin{table*}[ht!]
\centering
\resizebox{\textwidth}{!}
    {
    \setlength{\tabcolsep}{7pt}
    \begin{tabular}{clllrrrrrrrrrr}
        \toprule
        &
        \textbf{Setting} &
        \textbf{Model} &
        \textbf{Class} &
        \textbf{C. Rep.} &
        \textbf{I. Rep.} &
        \textbf{S. class F1} &
        \textbf{I. Cons.} &
        \textbf{E. Cons.} &
        \textbf{Extra} &
        \textbf{Missing} &
        \textbf{S3} &
        \textbf{S5} &
        \textbf{S7}\\ 
        \midrule
         \multirow{25}{*}{\rotatebox[origin=c]{90}{\textbf{Explanation}}} & \multirow{10}{*}{Zero-shot} &
          \multirow{5}{*}{GPT-4} & All & 44.23 & 17.31 & 78.79 & 00.19 & 00.21 &03.85 & 00.12 & 76.92 & 73.08 & 36.54 \\
           & & & True & 45.16 & 16.13 & 100.0 & 00.13 & 00.16 & 06.45 & 00.06 & 87.1 & 80.65 & 41.94

 \\
           & & & False & 50.00 & 14.29 & 88.00 & 00.14 & 00.21 & 00.00 & 00.07 & 71.43 & 71.43 & 35.71

 \\
           & & & Mixture & 50.00 & 25.00 & 85.71 & 00.50 & 00.75 & 00.00 & 00.50 & 50.0 & 50.0 & 0.0

  \\
           & & & Unproven & 00.00 & 33.33 & 80.00 & 00.67 & 00.00 & 00.00 & 00.33 & 33.33 & 33.33 & 33.33

 \\
            \cmidrule(lr){3-14}
           & & \multirow{5}{*}{Llama-70B} & All & 44.23 & 17.31 & 77.13 & 00.31 & 00.37 & 03.85 & 00.42 & 65.38 & 65.38 & 23.08 \\
           & & & True & 45.16 & 09.68 & 94.92 & 00.26 & 00.45 & 06.45 & 00.39 & 74.19 & 74.19 & 32.26

 \\
           & & & False & 42.86 & 28.57 & 83.33 & 00.57 & 00.36 & 00.00 & 00.50 & 57.14 & 57.14 & 14.29

 \\
           & & & Mixture & 25.00 & 25.00 & 85.71 &00.00 & 00.00 & 00.00 & 00.25 & 50.0 & 50.0 & 0.0

 \\
           & & & Unproven & 66.67 & 33.33 & 100.0 & 00.00 & 00.00 & 00.00 & 00.67 & 33.33 & 33.33 & 0.0

  \\
          \cmidrule(lr){2-14}  
        
          & \multirow{10}{*}{Few-shot} &
          \multirow{5}{*}{GPT-4} & All & 05.77 & 01.92 & 56.75 & 00.04 & 00.40 & 38.46 & 00.56 & 42.31 & 42.31 & \textbf{38.46} \\
          
           & & & True & 00.00 & 03.23 & 91.23 & 00.00 & 00.39 & 41.94 & 00.61 & 45.16 & 45.16 & 45.16

  \\
           & & & False & 14.29 & 00.00 & 72.73 & 00.07 & 00.43 & 28.57 & 00.57 & 42.86 & 42.86 & 35.71

 \\
           & & & Mixture & 00.00 & 00.00 & 40.00 & 00.00 & 00.00 & 25.00 & 00.00 & 25.0 & 25.0 & 25.0

 \\
           & & & Unproven & 33.33 & 00.00& 100.0 & 00.33 & 01.00 & 66.67 & 00.67 & 33.33 & 33.33 & 0.0

 \\
            \cmidrule(lr){3-14}
           & & \multirow{5}{*}{Llama-70B} & All & 30.77 & 01.92 & 45.68 & 00.04 & 00.31 & 05.77 & 00.96 & 32.69 & 32.69 & 25.0 \\
           & & & True & 25.81 & 03.23 & 87.27 & 00.00 & 00.13 & 06.45 & 01.03 & 35.48 & 35.48 & 29.03

  \\
           & & & False & 50.00 & 00.00 & 66.67 & 00.07 & 00.29 & 07.14 & 00.71 & 35.71 & 35.71 & 21.43

  \\
           & & & Mixture & 00.00 & 00.00 & 40.00 & 00.00 & 00.00 & 00.00 & 00.50 & 25.0 & 25.0 & 25.0

 \\
           & & & Unproven & 33.33 & 00.00 & 50.00 & 0.33 & 02.67 & 00.00 & 02.00 & 0.0 & 0.0 & 0.0

 \\
         \cmidrule(lr){2-14}

          & \multirow{5}{*}{PEFT} &
          \multirow{5}{*}{Vicuna-13B} & All & 23.08 & 09.62 & 50.61 & 00.04 & 00.25 & 25.00 & 00.63 & 36.54 & 36.54 & 25.0 \\
           & & & True & 16.13 & 09.68 & 93.10 & 00.00 & 00.10 & 29.03 & 00.48 & 45.16 & 45.16 & 35.48

  \\
           & & & False & 42.86 & 07.14 & 60.00 & 00.14 & 00.71 & 21.43 & 00.79 & 28.57 & 28.57 & 14.29
  \\
           & & & Mixture & 00.00 & 25.00 & 66.67 & 00.00 & 00.00 & 00.00 & 00.75 & 25.0 & 25.0 & 0.0
  \\
           & & & Unproven & 33.33 & 00.00 & 50.00 & 00.00 & 00.00 & 33.33 & 01.33 & 0.0 & 0.0 & 0.0
  \\
         \midrule
         
         \multirow{25}{*}{\rotatebox[origin=c]{90}{\textbf{Joint}}} & \multirow{10}{*}{Zero-shot} &
         \multirow{5}{*}{GPT-4} & All & 32.69 & 03.85 & 53.03 & 00.04 & 00.06 & 03.85 & 00.08 & 59.62 & 57.69 & 38.46 \\
           & & & True & 32.26 & 03.23 & 83.02 & 00.00 & 00.06 & 03.23 & 00.06 & 61.29 & 61.29 & 41.94
 \\
           & & & False & 35.71 & 07.14 & 83.33 & 00.07 & 00.00 & 07.14 & 00.00 & 71.43 & 71.43 & 42.86
 \\
           & & & Mixture & 25.00 & 00.00 & 40.00 & 00.00 & 00.00 & 00.00 & 00.00 & 25.0 & 25.0 & 25.0
  \\
           & & & Unproven & 33.33 & 00.00 & 80.00 & 00.33 & 00.33 & 00.00 & 00.67 & 33.33 & 0.0 & 0.0
  \\
            \cmidrule(lr){3-14}
           & & \multirow{5}{*}{Vicuna-13B} & All & 38.46 & 11.54 & 48.73 & 00.17 & 00.12 & 03.85 & 00.06 & 55.77 & 51.92 & 25.0 \\
           & & & True & 35.48 & 09.68 & 83.02 & 00.10 & 00.13 & 06.45 & 00.03 & 64.52 & 61.29 & 25.81
 \\
           & & & False & 50.00 & 14.29 & 60.00 & 00.43 & 00.14 & 00.00 & 00.07 & 42.86 & 35.71 & 21.43
  \\
           & & & Mixture & 00.00 & 00.00 & 85.71 & 00.00 & 00.00 & 00.00 & 00.25 & 50.0 & 50.0 & 50.0
 \\
           & & & Unproven & 66.67 & 33.33 & 50.0 & 00.00 & 00.00 & 00.00 & 00.00 & 33.33 & 33.33 & 0.0
  \\
          \cmidrule(lr){2-14}  
          
          & \multirow{10}{*}{Few-shot} &
          \multirow{5}{*}{GPT-3.5-D} & All & 03.85 & 00.00 & 50.52 & 00.00 & 00.06 & 00.00 & 00.46 & 51.92 & 51.92 & \textbf{48.08} \\
          
           & & & True & 06.45 & 00.00 & 93.10 & 00.00 & 00.10 & 00.00 & 00.35 & 61.29 & 61.29 & 54.84
 \\
           & & & False & 00.00 & 00.00 & 78.26 & 00.00 & 00.00 & 00.00 & 00.50 & 50.0 & 50.0 & 50.0
 \\
           & & & Mixture & 00.00 &00.00 & 00.00 & 00.00 & 00.00 & 00.00 & 01.00 & 0.0 & 0.0 & 0.0
 \\
           & & & Unproven & 00.00 & 00.00 & 80.00 & 00.00 & 00.00 & 00.00 & 00.67 & 33.33 & 33.33 & 33.33
  \\
            \cmidrule(lr){3-14}
           & & \multirow{5}{*}{Vicuna-13B} & All & 19.23 & 03.85 & 58.25 & 00.04 & 00.00 & 00.00 & 00.21 & 67.31 & 67.31 & \textbf{48.08}  \\
           
           & & & True & 12.90 & 03.23 & 93.10 & 00.03 & 00.00 & 00.00 & 00.23 & 74.19 & 74.19 & 61.29
  \\
           & & & False & 42.86 & 07.14 & 88.00 & 00.07 & 00.00 & 00.00 & 00.21 & 71.43 & 71.43 & 28.57
\\
           & & & Mixture & 00.00 & 00.00 & 00.00 & 00.00 & 00.00 & 00.00 & 00.00 & 0.0 & 0.0 & 0.0
  \\
           & & & Unproven & 00.00 & 00.00 & 100.0 & 00.00 & 00.00 & 00.00 & 00.33 & 66.67 & 66.67 & 66.67
 \\
         \cmidrule(lr){2-14}

        & \multirow{5}{*}{PEFT} &
          \multirow{5}{*}{Vicuna-13B} & All & 09.62 & 07.69 & 64.59 & 00.00 & 00.19 & 36.54 & 00.38 & 42.31 & 42.31 & 40.38 
          \\
           & & & True & 12.90 & 06.45 & 89.29 & 00.00 & 00.06 & 45.16 & 00.35 & 45.16 & 45.16 & 41.94
 \\
           & & & False & 07.14 & 07.14 & 78.26 & 00.00 & 00.36 & 21.43 & 00.36 & 42.86 & 42.86 & 42.86
  \\
           & & & Mixture & 00.00 & 25.00 & 85.71 & 00.00 & 00.25 & 25.00 & 00.50 & 25.0 & 25.0 & 25.0
 \\
           & & & Unproven & 00.00 & 00.00 & 80.00 & 00.00 & 00.67 & 33.33 & 00.67 & 33.33 & 33.33 & 33.33
  \\

  \cmidrule(lr){2-14}

        & \multirow{5}{*}{Gold Exp.} &
          \multirow{5}{*}{} & All & 07.69 & 00.00 & 41.72 & 00.08 & 00.17 & 50.00 & 00.48 & 25.00 & 25.00 & 19.23 
          \\
           & & & True & 06.45 & 00.00 & 89.29 & 00.00 & 00.03 & 64.52 & 00.32 & 25.81 & 25.81 & 19.35
 \\
           & & & False & 14.29 & 00.00 & 52.63 & 00.00 & 00.14 & 35.71 & 00.57 & 28.57 & 28.57 & 21.43
  \\
           & & & Mixture & 00.00 & 00.00 & 40.00 & 00.00 & 00.50 & 25.00 & 00.50 & 25.00 & 25.00 & 25.00
 \\
           & & & Unproven & 00.00 & 00.00 & 50.00 & 01.33 & 01.33 & 00.00 & 01.67 & 00.00 & 00.00 & 00.00
  \\
        \bottomrule
    \end{tabular}
    }
  \caption{Comprehensive human evaluation results for the best models, categorized by class: C. Rep. is Claim Repetition, I. Rep is Internal Repetition, S. Class F1 is the Suggested Class F1, I. Cons. is Internal Consistency, E. Cons. is External Consistency, Extra is Extra Information and Missing is Missing Information.}
  \label{detailed-human-evaluation-results} 
\end{table*}

\clearpage\begin{table*}[p]
\centering

\begin{subtable}[t]{\textwidth}
    \setlength{\tabcolsep}{9pt}
    \resizebox{\textwidth}{!}{
    \begin{tabular}{p{10em}p{2.5em}p{2.5em}p{2.5em}p{2.5em}p{2.5em}p{2.5em}p{2.5em}p{2.5em}p{2.5em}p{2.5em}p{2.5em}}
    \toprule
    \textbf{Model} & \rotatebox{90}{Zero-shot GPT-4 (E)} & \rotatebox{90}{Zero-shot Llama-70B (E)} & \rotatebox{90}{Few-shot GPT-4 (E)} & \rotatebox{90}{Few-shot Llama-70B (E)} & \rotatebox{90}{PEFT Vicuna-13B (E)} & \rotatebox{90}{Zero-shot GPT-4 (J)} & \rotatebox{90}{Zero-shot Vicuna-13B (J)} & \rotatebox{90}{Few-shot GPT-3.5-D (J)} & \rotatebox{90}{Few-shot Vicuna-13B (J)} & \rotatebox{90}{PEFT Vicuna-13B (J)} & \rotatebox{90}{Gold Explanation} \\
    \midrule
    Zero-shot GPT-4 (E) & - & $0.21$ & $0.00^*$ & $0.00^*$ & $0.00^*$ & $0.04^*$ & $0.01^*$ & $0.00^*$ & $0.30$ & $0.00^*$ & $0.00^*$ \\
    Zero-shot Llama-70B (E) & $0.21$ & - & $0.03^*$ & $0.00^*$ & $0.01^*$ & $0.63$ & $0.36$ & $0.19$ & $1.00$ & $0.03^*$ & $0.00^*$ \\
    Few-shot GPT-4 (E) & $0.00^*$ & $0.03^*$ & - & $0.39$ & $0.62$ & $0.10$ & $0.21$ & $0.40$ & $0.01^*$ & $1.00$ & $0.06$ \\
    Few-shot Llama-70B (E) & $0.00^*$ & $0.00^*$ & $0.39$ & - & $0.82$ & $0.01^*$ & $0.01^*$ & $0.02^*$ & $0.00^*$ & $0.44$ & $0.51$ \\
    PEFT Vicuna-13B (E) & $0.00^*$ & $0.01^*$ & $0.62$ & $0.82$ & - & $0.02^*$ & $0.05^*$ & $0.12$ & $0.00^*$ & $0.66$ & $0.28$ \\
    Zero-shot GPT-4 (J) & $0.04^*$ & $0.63$ & $0.10$ & $0.01^*$ & $0.02^*$ & - & $0.82$ & $0.49$ & $0.49$ & $0.11$ & $0.00^*$ \\
    Zero-shot Vicuna-13B (J) & $0.01^*$ & $0.36$ & $0.21$ & $0.01^*$ & $0.05^*$ & $0.82$ & - & $0.80$ & $0.27$ & $0.25$ & $0.00^*$ \\
    Few-shot GPT-3.5-D (J) & $0.00^*$ & $0.19$ & $0.40$ & $0.02^*$ & $0.12$ & $0.49$ & $0.80$ & - & $0.11$ & $0.40$ & $0.01^*$ \\
    Few-shot Vicuna-13B (J) & $0.30$ & $1.00$ & $0.01^*$ & $0.00^*$ & $0.00^*$ & $0.49$ & $0.27$ & $0.11$ & - & $0.02^*$ & $0.00^*$ \\
    PEFT Vicuna-13B (J) & $0.00^*$ & $0.03^*$ & $1.00$ & $0.44$ & $0.66$ & $0.11$ & $0.25$ & $0.40$ & $0.02^*$ & - & $0.06$ \\
    Gold Explanation & $0.00^*$ & $0.00^*$ & $0.06$ & $0.51$ & $0.28$ & $0.00^*$ & $0.00^*$ & $0.01^*$ & $0.00^*$ & $0.06$ & - \\
    \bottomrule
    \end{tabular}}
\caption{\footnotesize S3}
\end{subtable}\medskip

\begin{subtable}[t]{\textwidth}
    \setlength{\tabcolsep}{9pt}
    \resizebox{\textwidth}{!}{
    \begin{tabular}{p{10em}p{2.5em}p{2.5em}p{2.5em}p{2.5em}p{2.5em}p{2.5em}p{2.5em}p{2.5em}p{2.5em}p{2.5em}p{2.5em}}
    \toprule

    Zero-shot GPT-4 (E) & - & $0.48$ & $0.00^*$ & $0.00^*$ & $0.00^*$ & $0.10$ & $0.02^*$ & $0.01^*$ & $0.63$ & $0.00^*$ & $0.00^*$ \\
    Zero-shot Llama-70B (E) & $0.48$ & - & $0.03^*$ & $0.00^*$ & $0.01^*$ & $0.46$ & $0.17$ & $0.19$ & $1.00$ & $0.03^*$ & $0.00^*$ \\
    Few-shot GPT-4 (E) & $0.00^*$ & $0.03^*$ & - & $0.39$ & $0.62$ & $0.15$ & $0.41$ & $0.40$ & $0.01^*$ & $1.00$ & $0.06$ \\
    Few-shot Llama-70B (E) & $0.00^*$ & $0.00^*$ & $0.39$ & - & $0.82$ & $0.01^*$ & $0.03^*$ & $0.02^*$ & $0.00^*$ & $0.44$ & $0.51$ \\
    PEFT Vicuna-13B (E) & $0.00^*$ & $0.01^*$ & $0.62$ & $0.82$ & - & $0.03^*$ & $0.13$ & $0.12$ & $0.00^*$ & $0.66$ & $0.28$ \\
    Zero-shot GPT-4 (J) & $0.10$ & $0.46$ & $0.15$ & $0.01^*$ & $0.03^*$ & - & $0.67$ & $0.65$ & $0.37$ & $0.15$ & $0.00^*$ \\
    Zero-shot Vicuna-13B (J) & $0.02^*$ & $0.17$ & $0.41$ & $0.03^*$ & $0.13$ & $0.67$ & - & $1.00$ & $0.14$ & $0.47$ & $0.00^*$ \\
    Few-shot GPT-3.5-D (J) & $0.01^*$ & $0.19$ & $0.40$ & $0.02^*$ & $0.12$ & $0.65$ & $1.00$ & - & $0.11$ & $0.40$ & $0.01^*$ \\
    Few-shot Vicuna-13B (J) & $0.63$ & $1.00$ & $0.01^*$ & $0.00^*$ & $0.00^*$ & $0.37$ & $0.14$ & $0.11$ & - & $0.02^*$ & $0.00^*$ \\
    PEFT Vicuna-13B (J) & $0.00^*$ & $0.03^*$ & $1.00$ & $0.44$ & $0.66$ & $0.15$ & $0.47$ & $0.40$ & $0.02^*$ & - & $0.06$ \\
    Gold Explanation & $0.00^*$ & $0.00^*$ & $0.06$ & $0.51$ & $0.28$ & $0.00^*$ & $0.00^*$ & $0.01^*$ & $0.00^*$ & $0.06$ & - \\
    \bottomrule
    \end{tabular}}
\caption{\footnotesize S5}
\end{subtable}\medskip

\begin{subtable}[t]{\textwidth}
    \setlength{\tabcolsep}{9pt}
    \resizebox{\textwidth}{!}{
    \begin{tabular}{p{10em}p{2.5em}p{2.5em}p{2.5em}p{2.5em}p{2.5em}p{2.5em}p{2.5em}p{2.5em}p{2.5em}p{2.5em}p{2.5em}}
    \toprule

    Zero-shot GPT-4 (E) & - & $0.14$ & $1.00$ & $0.31$ & $0.31$ & $1.00$ & $0.28$ & $0.26$ & $0.33$ & $0.81$ & $0.06$ \\
    Zero-shot Llama-70B (E) & $0.14$ & - & $0.15$ & $1.00$ & $1.00$ & $0.12$ & $1.00$ & $0.01^*$ & $0.02^*$ & $0.09$ & $0.81$ \\
    Few-shot GPT-4 (E) & $1.00$ & $0.15$ & - & $0.16$ & $0.11$ & $1.00$ & $0.19$ & $0.42$ & $0.38$ & $1.00$ & $0.04^*$ \\
    Few-shot Llama-70B (E) & $0.31$ & $1.00$ & $0.16$ & - & $1.00$ & $0.22$ & $1.00$ & $0.01^*$ & $0.02^*$ & $0.16$ & $0.60$ \\
    PEFT Vicuna-13B (E) & $0.31$ & $1.00$ & $0.11$ & $1.00$ & - & $0.15$ & $1.00$ & $0.02^*$ & $0.00^*$ & $0.12$ & $0.63$ \\
    Zero-shot GPT-4 (J) & $1.00$ & $0.12$ & $1.00$ & $0.22$ & $0.15$ & - & $0.19$ & $0.41$ & $0.47$ & $1.00$ & $0.05$ \\
    Zero-shot Vicuna-13B (J) & $0.28$ & $1.00$ & $0.19$ & $1.00$ & $1.00$ & $0.19$ & - & $0.02^*$ & $0.03^*$ & $0.15$ & $0.63$ \\
    Few-shot GPT-3.5-D (J) & $0.26$ & $0.01^*$ & $0.42$ & $0.01^*$ & $0.02^*$ & $0.41$ & $0.02^*$ & - & $1.00$ & $0.55$ & $0.00^*$ \\
    Few-shot Vicuna-13B (J) & $0.33$ & $0.02^*$ & $0.38$ & $0.02^*$ & $0.00^*$ & $0.47$ & $0.03^*$ & $1.00$ & - & $0.53$ & $0.01^*$ \\
    PEFT Vicuna-13B (J) & $0.81$ & $0.09$ & $1.00$ & $0.16$ & $0.12$ & $1.00$ & $0.15$ & $0.55$ & $0.53$ & - & $0.02^*$ \\
    Gold Explanation & $0.06$ & $0.81$ & $0.04^*$ & $0.60$ & $0.63$ & $0.05$ & $0.63$ & $0.00^*$ & $0.01^*$ & $0.02^*$ & - \\
    \bottomrule
    \end{tabular}}
\caption{\footnotesize S7}
\end{subtable}\medskip

\caption{Results of a paired two-sided randomization test (10'000 rounds) on the human evaluation results with $\alpha = 0.05$. The upper half indicates $p$-values for different human evaluation criteria. Parentheses in model names indicate the task, where (E) stands for an explanation only model and (J) for a joint model.}
\label{human_eval-stest}
\end{table*}\clearpage

\begin{figure*}
  \begin{subfigure}{.5\textwidth}
  \centering
    \includegraphics[width=.9\linewidth]{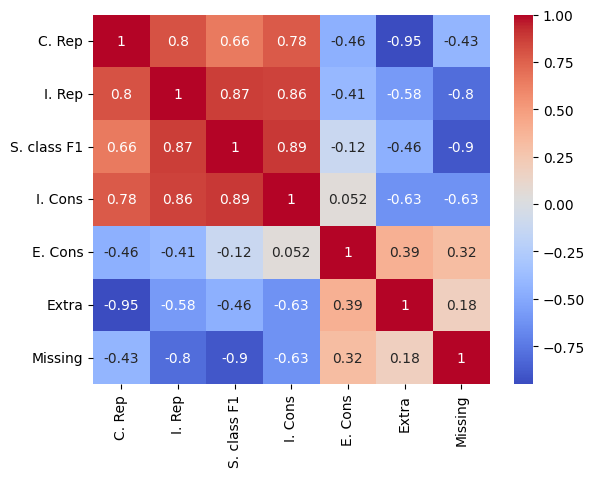}
    \caption{Explanation}
  \end{subfigure}%
  \begin{subfigure}{.5\textwidth}
  \centering
    \includegraphics[width=.9\linewidth]{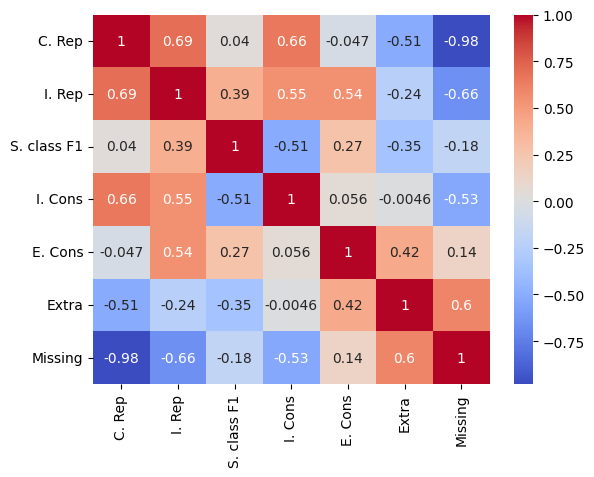}
    \caption{Joint}
  \end{subfigure}
  \caption{The correlation between different human evaluation metrics for each task.}
  \label{merged_heatmap}
\end{figure*}

\end{document}